\documentclass[sigconf,preprint]{acmart}
\AtBeginDocument{%
  }

\setcopyright{acmlicensed}
\copyrightyear{2018}
\acmYear{2018}
\acmDOI{XXXXXXX.XXXXXXX}
\acmConference[Conference acronym 'XX]{Make sure to enter the correct
  conference title from your rights confirmation email}{June 03--05,
  2018}{Woodstock, NY}
\acmISBN{978-1-4503-XXXX-X/2018/06}

\usepackage{amsmath}
\usepackage{amsthm}
\usepackage{booktabs}
\usepackage{framed} 
\usepackage{multirow}
\usepackage{enumitem}
\usepackage{caption}
\usepackage{graphicx}
\usepackage{float} 
\usepackage{etoolbox}
\usepackage{xspace}
\usepackage{tabularx}
\usepackage{tcolorbox}
\tcbuselibrary{skins,breakable}
\usepackage{accents}
\usepackage{subfig}
\usepackage{makecell}
\usepackage{bbding}
\usepackage{multicol}
\usepackage[table]{xcolor}
\usepackage{hyperref}
\usepackage[normalem]{ulem}

\newcommand{\eg}{\emph{e.g.},\xspace}

\newcommand{\etc}{\emph{etc.}\xspace}
\newcommand{\model}{SIREN}
\newcommand{\bench}{\model-Bench}

\newcommand\figref[1]{Figure~\ref{#1}}

\newcommand\tabref[1]{Table~\ref{#1}}
\newcommand\secref[1]{Section~\ref{#1}}

\newcommand\appref[1]{Appendix~\ref{#1}}

\newcommand{\eat}[1]{}

\newcommand{\jia}[1]{{\color{purple}{#1}}}

\begin{document}

\title{\model: Towards End-to-End Extreme-Weather Early Warning with Experience-Grounded LLM Agents}

\author{Hang Ni}
\affiliation{%
\department{Artificial Intelligence Thrust}
  \institution{The Hong Kong University of Science and Technology (Guangzhou)}
  \city{Guangzhou}
  \country{China}}
\email{hni017@connect.hkust-gz.edu.cn}

\author{Weijia Zhang}
\affiliation{%
\department{Artificial Intelligence Thrust}
  \institution{The Hong Kong University of Science and Technology (Guangzhou)}
  \city{Guangzhou}
  \country{China}
  }
\email{wzhang411@connect.hkust-gz.edu.cn}

\author{Fan Liu}
\affiliation{%
\department{Artificial Intelligence Thrust}
  \institution{The Hong Kong University of Science and Technology (Guangzhou)}
  \city{Guangzhou}
  \country{China}
  }
\email{fliu236@connect.hkust-gz.edu.cn}

\author{Mengqian Lu}
\affiliation{%
    \department{Department of Civil and Environmental Engineering}
  \institution{The Hong Kong University of Science and Technology}
  \city{Hong Kong}
  \country{China}
  }
\additionalaffiliation{Otto Poon Center for Climate Resilience and Sustainability, The World Sustainable Development Institute}
\email{mengqian.lu@ust.hk}

\author{Hao Liu}
\affiliation{%
    \department{Artificial Intelligence Thrust}
  \institution{The Hong Kong University of Science and Technology (Guangzhou)}
  \city{Guangzhou}
  \country{China}
  }
\email{liuh@ust.hk}

\renewcommand{\shortauthors}{Hang Ni, Weijia Zhang, Fan Liu, Mengqian Lu, Hao Liu}

\begin{abstract}
Early warning of extreme weather is essential for mitigating the societal, economic, and environmental risks posed by hazardous weather events.
However, expert-centered warning workflows are costly, labor-intensive, and difficult to scale throughout the warning-to-action process.
Although recent advances in Large Language Model (LLM) agents have enabled the automation of weather-related tasks, existing studies remain centered on isolated scientific tasks and overlook the chain of interdependent processes required for operational extreme-weather early warning.
To bridge this gap, this study investigates automated end-to-end extreme-weather early warning through LLM agents.
We first develop \textbf{\bench}, a comprehensive benchmark comprising 600 question-answer instances across 19 tasks, and covering four individual warning procedures and an end-to-end warning chain.
Evaluation on \bench\xspace reveals substantial capability gaps in existing weather agent frameworks. 
This motivates us to develop \textbf{\model}, an experience-grounded agent framework inspired by experts' use of historical cases, which combines an agentic execution environment integrating heterogeneous weather evidence and tools with a family of agent harnesses that exploit historical cases through retrieval, skill distillation, and predictive modeling.
Extensive experiments demonstrate that \model\xspace outperforms weather-agent baselines on both individual warning procedures and end-to-end warning chains.
\end{abstract}

\eat{\jia{Although recent advances in large language model (LLM) agents have enabled the automation of a growing range of weather-related tasks, existing studies remain centered on isolated scientific tasks, such as hazard analysis and prediction, and overlook the chain of interdependent processes required for operational extreme-weather early warning.
To bridge this gap, this study investigates end-to-end automated extreme-weather early warning through LLM agents.
We first develop \textbf{\bench}, a comprehensive benchmark for extreme-weather early warning, which comprises 600 question-answer instances across 19 tasks, covering four individual warning procedures and an end-to-end warning chain.
Evaluation on \bench\xspace reveals substantial capability gaps in existing automated weather frameworks. This motivates us to develop \textbf{\model}, an experience-grounded agent framework inspired by experts' use of historical cases, which combines an agentic execution environment integrating heterogeneous weather evidence and tools with a family of agent harnesses that exploit historical cases through retrieval, skill distillation, and predictive modeling.
Extensive experiments demonstrate that \model\xspace outperforms weather-agent baselines on both individual warning procedures and end-to-end warning chains.}}




\maketitle

\section{Introduction}

The increasing frequency, intensity, and compound impacts of extreme weather events are amplifying systemic risks to human societies, economic activities, and ecosystems~\cite{camps2025artificial}.
In response, the United Nations (UN) and the World Meteorological Organization (WMO) have identified Early Warning Systems (EWS) as a cornerstone of disaster risk reduction and climate adaptation~\cite{un_ew4all,wmo_ew4all}.
An effective EWS is not merely a forecasting module but an operational warning-to-action chain that integrates hazard monitoring, event localization, impact-based risk assessment, warning communication, and preparedness decision-making~\cite{reichstein2025early}.
Its objective is therefore to move beyond hazard-centered prediction toward actionable, impact-oriented warnings that enable timely interventions before hazardous weather develops into a severe disaster~\cite{de2022learning}.

Implementing such an integrated warning pipeline remains challenging.
In current practice, extreme-weather warning relies heavily on domain experts to interpret evolving atmospheric conditions, apply meteorological knowledge and prior experience, coordinate heterogeneous data sources and analytical tools, and synthesize intermediate evidence into warning and response decisions~\cite{wmo2015impact,potter2021benefits}.
Despite its effectiveness, this expert-centered workflow is costly, labor-intensive, difficult to scale, and vulnerable to cascading errors across interdependent stages.
Existing software systems and machine learning (ML) models have improved individual components, such as detection, forecasting, and diagnostic analysis, but typically address isolated tasks, leaving human experts to integrate evidence and manage the end-to-end warning process.
Recent advances in LLMs have enabled a new paradigm of data-driven LLM agents~\cite{DataInterpreter2025,liu2026towards}, which can automatically orchestrate multi-step reasoning, generate code, and use scientific tools to transform heterogeneous evidence into grounded analyses.
These advances motivate us to formulate extreme-weather early warning as an agentic task in which LLM agents gather meteorological evidence, reason about hazard risks, and support operational decisions.

\begin{figure}[tbp]
    \centering
    \includegraphics[width=\linewidth]{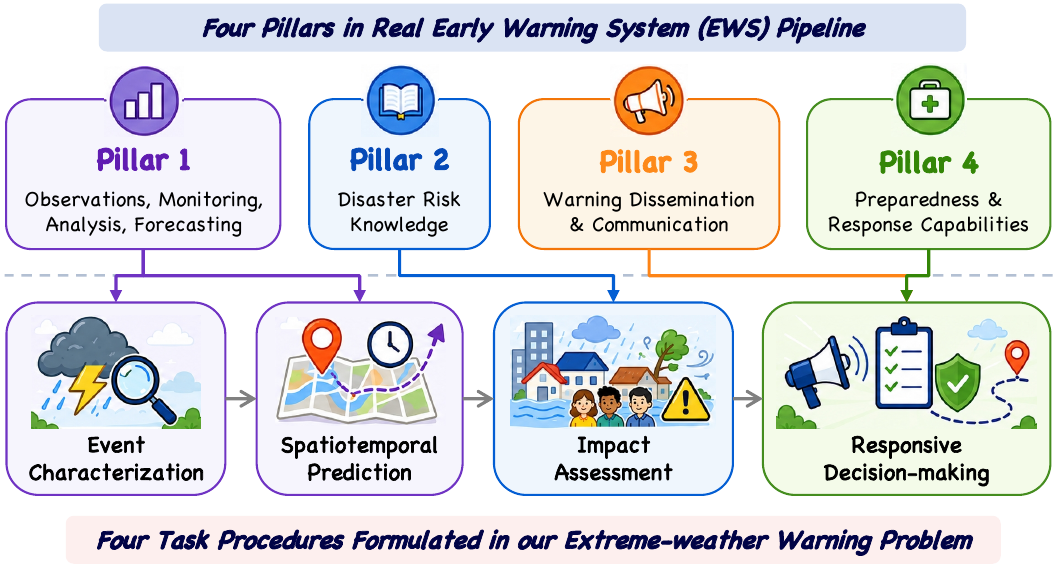}
    \vspace{-10pt}
    \caption{Connections to the EWS pipeline~\cite{reichstein2025early}.}
    \vspace{-10pt}
    \label{fig:motivation}
\end{figure}

Recent studies have investigated LLM agents for data-driven weather applications, including extreme-weather diagnosis~\cite{jiang2025ewe,hvrmet2026} and automated weather-science workflows~\cite{guo2025self,kim2025climateagent,zephyrus2026,climagent2026,zhang2026tianji}.
However, these agents primarily focus on early-stage tasks, such as hazard analysis and forecasting, while largely omitting downstream impact assessment, warning communication, and response-oriented decision support.
Consequently, existing benchmarks do not capture the full operational objectives of end-to-end extreme-weather warning.
To address this gap, we construct \textbf{\bench}\footnote{The benchmark is available at \url{https://anonymous.4open.science/r/SIREN-5CE8/}.}, a comprehensive benchmark based on over 12 types of extreme-weather events across the U.S. region in 2021, comprising 600 question-answer (QA) pairs.
As shown in \figref{fig:motivation}, \bench\xspace follows the warning-to-action perspective of EWS~\cite{reichstein2025early} and organizes extreme-weather warning into four core task procedures: \textit{event characterization} to interpret ongoing hazards, \textit{spatiotemporal prediction} to forecast their evolution, \textit{impact assessment} to estimate potential consequences, and \textit{responsive decision-making} to translate evidence into warnings and interventions.
It further integrates these individual capabilities into a \textit{warning chain} task that requires agents to orchestrate the end-to-end operational pipeline.
As summarized in \tabref{tab:benchmark_coverage}, \bench\xspace provides the broadest coverage by spanning all five task categories.
Nevertheless, evaluations of five recent weather-agent frameworks reveal substantial capability gaps, indicating that existing agents remain far from operational readiness for both individual procedures and warning chains.

\newcommand{\PartialCoverageMark}{%
  \tikz[baseline=(partialcheck.base)]{%
    \node[inner sep=0pt,outer sep=0pt] (partialcheck) {\CheckmarkBold};
    \draw[line width=0.85pt,line cap=round]
      ([xshift=-3.2pt,yshift=-0.2pt]partialcheck.north east) --
      ([xshift=-0.3pt,yshift=-3.1pt]partialcheck.north east);
    \draw[line width=0.85pt,line cap=round]
      ([xshift=-0.3pt,yshift=-0.2pt]partialcheck.north east) --
      ([xshift=-3.2pt,yshift=-3.1pt]partialcheck.north east);%
  }%
}
\begin{table}[t]
\centering
\caption{Comparisons of benchmark coverage, with EC, SP, IA, RD, and WC denoting five task categories, respectively.}
\vspace{-5pt}
\label{tab:benchmark_coverage}
\small
\setlength{\tabcolsep}{3.5pt}
\renewcommand{\arraystretch}{1.08}
\begin{tabular}{l|cccc|c}
\toprule
\textbf{Benchmark} & \textbf{EC} & \textbf{SP} & \textbf{IA} & \textbf{RD} & \textbf{WC} \\
\midrule
ZephyrusBench~\cite{zephyrus2026} & \CheckmarkBold & \CheckmarkBold & \XSolidBrush & \XSolidBrush & \XSolidBrush \\
Jiang et al.~\cite{jiang2025ewe} & \CheckmarkBold & \XSolidBrush & \XSolidBrush & \XSolidBrush & \XSolidBrush \\
ClimaBench~\cite{climagent2026} & \PartialCoverageMark & \CheckmarkBold & \XSolidBrush & \PartialCoverageMark & \XSolidBrush \\
Climate-Agent-Bench-85~\cite{kim2025climateagent} & \CheckmarkBold & \CheckmarkBold & \XSolidBrush & \XSolidBrush & \XSolidBrush \\
Tang et al.~\cite{hvrmet2026} & \CheckmarkBold & \XSolidBrush & \XSolidBrush & \XSolidBrush & \XSolidBrush \\
\midrule
\textbf{\bench} & \CheckmarkBold & \CheckmarkBold & \CheckmarkBold & \CheckmarkBold & \CheckmarkBold \\
\bottomrule
\end{tabular}

\footnotesize \CheckmarkBold: covered; \PartialCoverageMark: partially covered; \XSolidBrush: not covered.
\vspace{-10pt}
\end{table}

In practice, extreme-weather analysis and warning decisions are rarely made in isolation.
For example, estimating injuries caused by a tornado requires not only analysis of the physical mechanisms of the ongoing event but also reference to prior cases with similar spatiotemporal or hazard characteristics.
Existing weather agents primarily reason over the current meteorological state while underusing historical cases that can provide reusable operational experience through retrieval, adaptation, or task-specific distillation.
To address this limitation, we introduce \textbf{\model}, an experience-grounded agent framework for operational extreme-weather early warning.
Specifically, \model\xspace is built on a comprehensive and executable \textit{agentic environment} that provides diverse weather-oriented analytical tools and interfaces for multimodal evidence sources, including meteorological grids, mesoanalysis images, geospatial information, \etc, thereby enabling LLM agents to ground their analyses and decisions in verifiable evidence rather than relying solely on parametric knowledge.
To incorporate historical operational experience, we construct a historical case database from question-answer pairs collected in earlier years and design \textit{a family of agents instantiated with diverse harnesses} that capture complementary uses of historical cases in agentic reasoning~\cite{dsagent2024,yang2026evods,mlebench2025}:
(1) \textit{\model-Base} does not use historical cases;
(2) \textit{\model-RAG} retrieves similar historical cases as direct analogical references for the current task;
(3) \textit{\model-Skill} performs deliberate rehearsal on informative cases and distills reusable procedural skills from their solution trajectories; and
(4) \textit{\model-Modeling} derives task-specific predictive indicators, constructs training and validation sets from the complete historical database, and develops ML models to support downstream analyses.
Extensive experiments demonstrate that grounding agents in historical experience substantially improves operational performance on both individual procedures and end-to-end warning chains.

Our main contributions are summarized as follows:
(1) To the best of our knowledge, \bench\xspace is the first benchmark for end-to-end extreme-weather early warning, comprising 600 QA pairs across 12 types of extreme events;
(2) \model\xspace is the first agent framework for end-to-end extreme-weather early warning, consisting of an agentic environment that integrates heterogeneous weather evidence and analytical tools, and a family of agents that leverage historical cases through complementary experience-grounded harnesses; and
(3) extensive experiments demonstrate strong performance against competitive weather-agent baselines on both individual procedures and operational warning chains.
\section{\bench}
This section presents \bench, a benchmark for evaluating LLM agents in operational extreme-weather early warning.

\subsection{Task Formulation}
Each instance in \bench\xspace is defined as a tuple $\tau=(q, y)$, where $q$ denotes an extreme-weather analysis question and $y$ denotes the ground-truth answer or reference response.
Given $q$, an LLM agent with policy $\pi_{\theta}$ produces an answer $\hat{y}=\pi_{\theta}(q)$ by retrieving evidence, invoking tools, reasoning over intermediate results, and synthesizing a final response.
The objective is to generate an answer $\hat{y}$ that is accurate, grounded in evidence, and operationally useful.

\subsection{Benchmark Overview}
\subsubsection{Overall Coverage}
\bench\xspace focuses on extreme-weather events in the U.S. in 2021.
It contains 600 QA instances across five categories of extreme-weather analysis aligned with the EWS pipeline, covering 19 subtasks.
These instances span 12 primary hazard families, including convective storms, tropical systems, tornadoes, floods, winter weather, high-wind events, visibility hazards, heat, fire, drought, marine hazards, and cold-weather events.
They are also distributed across all months of the year, enabling the benchmark to capture seasonal variation in weather analysis.

\subsubsection{Task Categories}
Following the operational EWS framework \cite{reichstein2025early}, our dataset is organized into five task categories:
(1) \textit{event characterization} analyzes the nature of an ongoing hazard, including hazard type classification and event mechanism analysis, with 64 instances;
(2) \textit{spatiotemporal prediction} estimates where, when, and how an event will develop, including spatial localization and forecasts of event occurrence, severity, evolution, and duration, with 160 instances;
(3) \textit{impact assessment} evaluates potential event consequences, including the risks of injury, property damage, agricultural loss, and power outages, with 192 instances;
(4) \textit{responsive decision-making} converts analytical evidence into operational actions, \eg warning decisions, alert generation, hazard mitigation, and disaster-recovery site selection, with 160 instances;
and (5) \textit{warning chain} integrates the preceding categories into an end-to-end pipeline that reflects operational early-warning processes, with 24 instances.
The first four task categories comprise 18 types of atomic subtasks, as summarized in \appref{app:task_taxonomy}, while the fifth category, \textit{warning chain}, constitutes a single end-to-end subtask, yielding 19 atomic subtasks in total.

\subsection{Dataset Construction}
\bench\xspace is constructed through a four-stage pipeline: (1) raw data curation, (2) multi-source data preprocessing and alignment, (3) task instance generation, and (4) quality control.

\subsubsection{Raw Data Curation} 
Our task instances are derived from three U.S. data sources related to extreme weather:
(1) \textit{National Oceanic and Atmospheric Administration (NOAA) Databases}~\cite{noaa}:
The NOAA National Centers for Environmental Information (NCEI) maintains the Storm Events Database (SED)~\cite{noaa_sed}, which records various types of severe weather together with spatiotemporal information, event types (\eg tornadoes and heatwaves), and impacts (\eg numbers of injuries and deaths).
These metadata support tasks such as extreme-weather type classification, spatiotemporal prediction, and impact assessment.
In addition, the NOAA Storm Prediction Center (SPC) provides Mesoscale Discussions (MDs)~\cite{spc_md}, which are specialized meteorological statements issued by forecasters to describe mesoscale conditions associated with hazardous weather.
We collect these statements for extreme-weather mechanism analysis.
(2) \textit{Federal Emergency Management Agency (FEMA) Databases}~\cite{openfema}:
FEMA provides datasets on disaster impacts and response decisions.
We select records associated with extreme weather events for task construction.
Specifically, these datasets support property risk assessment, public warning generation, and hazard mitigation decision-making.
(3) \textit{Open Energy Data Initiative (OEDI)}~\cite{oedi}:
OEDI provides utility outage information resources~\cite{oedi_utility_outage} that link utility service areas to public outage information channels maintained by electric utilities.
We use entries associated with extreme weather to construct the outage duration assessment task.

\subsubsection{Multi-source Data Preprocessing and Alignment}
The preprocessing pipeline first filters event records containing missing or anomalous entries to obtain a consistent subset of the raw data.
We then map key metadata (\eg timestamps, spatial identifiers, event types, \etc) from different sources to a shared event schema because these sources use heterogeneous field definitions, value formats, units, and naming conventions.
This step retains source-specific details while normalizing the data for cross-source use.

Constructing warning chain tasks requires identifying records that refer to the same extreme event across data sources and organizing them into four sequential task procedures.
However, the spatiotemporal attributes of the same event often differ across sources.
For example, SED may record a tornado as occurring on February 9, 2021, at 5:20 PM, whereas FEMA records the same event at 4:50 PM, making cross-source alignment nontrivial.
To address this issue, we design a soft alignment strategy that matches events of the same type while allowing controlled spatiotemporal tolerance.
For example, two events are considered matched when the difference between their recorded timestamps is within a predefined threshold or when their recorded locations are in neighboring counties.

\subsubsection{Task Instance Generation}
Based on the normalized task metadata, we construct formal QA instances in which each question contains three components: (1) \textit{task instruction}, (2) \textit{event conditions}, and (3) \textit{answer specification}.
The task instruction defines the objective of the subtask, \eg identifying an event type or predicting an affected area, without prescribing an overly specific solution.
The event conditions provide the event-specific inputs required to complete the task, such as the reference timestamp, affected regions, and event descriptions, while excluding ground-truth information to prevent label leakage.
The answer specification constrains the output space according to the task requirements, \eg categorical labels, numerical values, or open-ended responses.
For chained tasks, the four question templates are arranged sequentially, and the final response combines the outputs of all four procedures for end-to-end evaluation.
This design better reflects operational chains, in which an agent must analyze the same event context through multiple cascading tasks rather than solve isolated problems.

\subsubsection{Quality Control}
To ensure benchmark quality, we audit the generated instances from 2021 through a three-stage process involving LLMs and human experts.
First, LLMs audit the instances according to predefined criteria: (1) whether the event conditions are incomplete or prone to leakage and (2) whether the ground-truth answers or reference responses are implausible, ambiguous, or insufficiently informative.
Low-quality instances are then discarded.
Subsequently, to prevent frequent hazards or disaster-prone months from dominating the benchmark, we resample the filtered instances to promote diversity and balanced coverage across extreme event types and occurrence seasons.
Finally, experts manually inspect the remaining instances to verify that they are plausible and representative of operational extreme-weather scenarios.

\subsection{Evaluation Protocols}
For task evaluation, we categorize task instances into four types based on their answer formats.

\noindent\textbf{Multiple-choice Classification (MC).} 
MC tasks require the agent to select one or more labels from a closed option set, such as hazard types or hazard mitigation actions. We report the accuracy measure based on exact label matching. 

\noindent\textbf{Numeric Regression (NR).} 
NR tasks require a scalar value, such as event duration or societal impact. 
Given a prediction $\hat{y}$ and reference $y$, we first compute the relative error as $\mathrm{RE}=|\hat{y}-y|/|y|$. 
We then report the bounded relative score $\mathrm{RS}=1/(1+\mathrm{RE})$, where higher values indicate better numerical agreement.

\noindent\textbf{Geospatial Localization (GL).} 
GL tasks evaluate open-form spatial answers, \eg affected administrative areas and recovery-center addresses. 
For state-county localization, we use an adjacency-aware score that assigns partial credit to nearby locations: $\mathrm{LS}=\lambda/(\delta_{\mathrm{state}}+1)+(1-\lambda)/(\delta_{\mathrm{county}}+1)$, where $\lambda$ controls the credit assigned to state matching, and $\delta_{\mathrm{state}}$ and $\delta_{\mathrm{county}}$ denote the minimum hop distances in the state- and county-level adjacency graphs, respectively. 
For coordinate or address localization, we geocode the prediction and reference when necessary, compute the minimum haversine distance $d$ in kilometers between the predicted and reference points, and report $\mathrm{SS}=1/(1+d)$ together with the best-distance statistic. 

\noindent\textbf{Open-ended Generation (OG).} 
OG tasks require free-form operational text, such as descriptions of physical mechanisms, public-warning instructions, or hazard mitigation responses. 
For most OG tasks, we extract a small set of essential points from the reference response and compute answer recall as $\mathrm{Recall}=|\mathcal{P}_{\mathrm{hit}}|/|\mathcal{P}_{\mathrm{ref}}|$, where $\mathcal{P}_{\mathrm{ref}}$ is the set of essential reference points and $\mathcal{P}_{\mathrm{hit}}$ is the subset supported by the agent response. 
For physical-understanding tasks, which require more holistic judgment, we use an expert-style LLM evaluator to score the answer against the reference on a 0--10 scale, with validation of LLM--human alignment provided in \appref{app:human_evaluation}.

For chained tasks, we do not assign a single monolithic metric. 
Instead, we parse the response into procedure-specific answers and evaluate each procedure using the corresponding MC, NR, GL, or OG metric. 
In addition to task-effectiveness metrics, we report two process-level measures for LLM agents: the completion pass rate $\mathrm{CPR}$, which indicates whether the agent successfully completes the task with a valid answer, and the execution pass rate $\mathrm{EPR}$, which indicates whether the agent completes the task without code-execution or post-check failures.

\begin{figure*}[ht]
    \centering
    \includegraphics[width=0.9\linewidth]{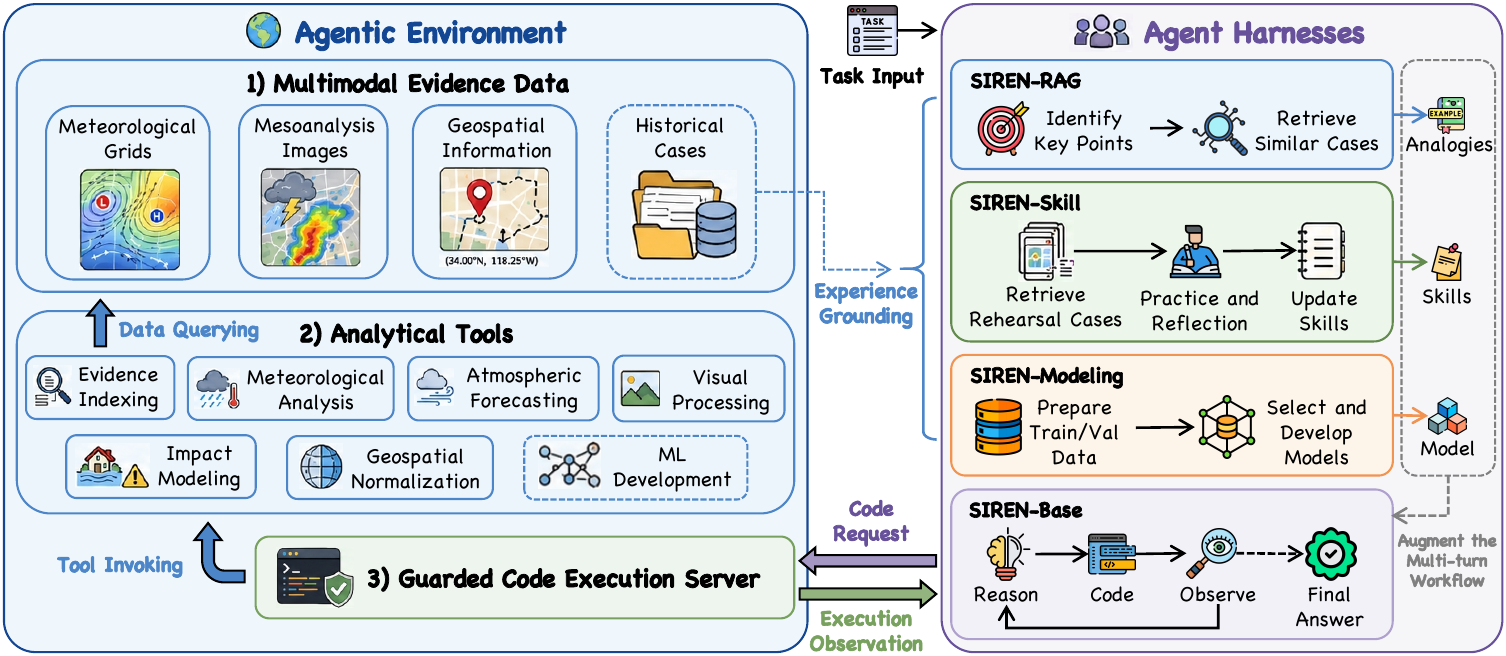}
    \vspace{-5pt}
    \caption{The overall framework of \model, which comprises an executable agentic environment and a family of agents instantiated by diverse experience-grounded harnesses.}
    \vspace{-5pt}
    \label{fig:main}
\end{figure*}

\section{\model}
This section presents \model, an agent framework for extreme-weather early warning. As illustrated in \figref{fig:main}, the framework comprises an executable agentic environment and a family of agents whose harnesses integrate historical cases through complementary experience-grounded reasoning mechanisms.

\subsection{Agentic Environment}
Operational early warning requires agents to track an evolving hazard from meteorological evidence to local impacts and response actions across interdependent stages, but existing weather-agent environments provide task-specific data and tools for isolated analysis or forecasting rather than an end-to-end workflow.
We therefore construct a unified agentic environment spanning these operational procedures, denoted by $\mathcal{E}$, which comprises multimodal evidence $\mathcal{X}$, analytical tools $\mathcal{T}$, and a guarded code execution server.

\subsubsection{Multimodal Evidence Data}
The environment provides access to four types of multimodal evidence $\mathcal{X}$:
(1) \textit{Meteorological grids} comprise high-resolution numerical weather analysis and forecast fields over the contiguous U.S., derived from the NOAA High-Resolution Rapid Refresh system~\cite{hrrr}.
They provide quantitative atmospheric states (\eg wind, temperature, and pressure) that serve as primary evidence for diagnostic reasoning and weather-field visualization.
(2) \textit{Mesoanalysis images} are operational severe-weather analysis graphics from the NOAA Storm Prediction Center Mesoanalysis archive~\cite{spc_mesoanalysis}, which summarizes surface objective analyses and short-range model fields as diagnostic maps for forecasters.
They complement gridded variables by presenting mesoscale ingredients, instability patterns, and storm-environment structures in a form suitable for visual inspection by agents.
(3) \textit{Geospatial information} combines U.S. administrative boundary files from the Census Bureau TIGER/Line products~\cite{census_tiger} with map features from OpenStreetMap~\cite{openstreetmap}.
It anchors meteorological evidence to counties, populated places, transportation networks, and local facilities, thereby supporting localization and exposure-aware reasoning.
(4) \textit{Historical cases} constitute a case base $\mathcal{K}=\{\mathcal{K}_k\}_{k=1}^{K}$, where each case $\mathcal{K}_k=(q_k,y_k)$ contains a warning question and the corresponding reference answer from a previous year.
The event conditions associated with $q_k$ are retained in the question context, allowing agents to compare the current task with prior operational situations.

\subsubsection{Analytical Tools}
The environment exposes analytical tools $\mathcal{T}$ in seven high-level categories:
(1) \textit{Evidence indexing} provides controlled access to evidence sources, allowing agents to retrieve task-relevant data while maintaining a clear separation between reasoning and data storage;
(2) \textit{Meteorological analysis} derives physical diagnostics from weather fields (\eg wind structures and thermodynamic summaries), converting raw variables into interpretable evidence;
(3) \textit{Atmospheric forecasting} provides access to advanced forecasting models, enabling agents to reason about future hazard evolution for operational decision-making;
(4) \textit{Visual processing} renders maps and gridded fields and supports lightweight image operations such as cropping, zooming, and annotation, enabling visual evidence to be inspected at the spatial scale required by each task;
(5) \textit{Impact modeling} estimates potential consequences from hazard intensity and exposed local assets, linking meteorological signals to operational risk assessment;
(6) \textit{Geospatial normalization} resolves place names, administrative units, and coordinates within a shared spatial frame, ensuring consistent location-based reasoning across weather, map, and benchmark records;
(7) \textit{ML development} provides interfaces for training task-specific predictive models using training and validation samples constructed by agents.
This category is enabled only for \model-Modeling, as discussed in \secref{sec:modeling-agent}.
Detailed tool descriptions are provided in \appref{app:analytical_tools}.

\subsubsection{Code Execution Server}
LLM agents require executable computation to analyze large-scale evidence, but generated code should not directly access raw storage paths or tool implementations.
We therefore route every Python snippet through a guarded code execution server that validates tool calls and path access, executes only approved operations in an isolated workspace, and returns structured observations to the agent. This design provides a controlled interface for data-intensive reasoning.

\subsection{Agent Harness}
We design a family of LLM agents that use $\mathcal{E}$ to solve operational extreme-weather warning tasks.
The \textit{\model-Base} agent follows a multi-turn code-execution workflow that alternates between code generation and observation-based reasoning before synthesizing the final solution.
However, this agent relies solely on current meteorological conditions and cannot use precedents from historical cases with similar spatiotemporal contexts.
To incorporate prior experience, we build three multi-stage harnesses on the base workflow, each corresponding to a representative strategy in recent data-driven agent research: case-level analogy~\cite{dsagent2024}, procedure-level skill accumulation~\cite{yang2026evods}, and data-level ML modeling~\cite{mlebench2025}. These strategies yield \textit{\model-RAG}, \textit{\model-Skill}, and \textit{\model-Modeling}, respectively.

\subsubsection{\model-Base}
\model-Base disables access to the historical case base $\mathcal{K}$ and relies only on the current question and event context.
Formally, the agent $\pi_{\theta}$ follows a multi-turn workflow:
\begin{equation}
\{r_t,a_t\}\sim \pi_{\theta}(\cdot \mid q, h_t),\quad
o_t=\mathcal{E}(a_t),\quad
h_{t+1}=h_t \cup \{r_t,a_t,o_t\}.
\end{equation}
At each turn $t\in\{1,\dots,T\}$, the agent produces a reasoning trace $r_t$ and executable code $a_t$ based on the current trajectory $h_t$ and then receives an execution observation $o_t$ from the environment $\mathcal{E}$.
The trajectory $h_t$ stores the intermediate outputs and observations accumulated up to turn $t$.
After the interaction terminates, the agent produces the final task solution as $\hat{y}=\pi_{\theta}(q,h_T)$.

\subsubsection{\model-RAG}
\model-RAG augments \model-Base with a retrieval harness for case-based analogy.
Guided by a retrieval instruction $I_{\mathrm{rag}}$, the agent first runs the code-execution loop with policy $\pi_{\theta}(\cdot\mid q,h_t,I_{\mathrm{rag}})$ to search $\mathcal{K}$ and identify a small set of cases $\mathcal{K}_{\mathrm{rag}}\subset\mathcal{K}$ that provide useful analogies for the current task based on both task intent and event conditions.
The retrieval phase terminates when the observation $o_{T_{\mathrm{rag}}}$ at turn $T_{\mathrm{rag}}$ returns $\mathcal{K}_{\mathrm{rag}}$.
The agent then resumes the base workflow with these cases as in-context references, following $\{r_t,a_t\}\sim\pi_{\theta}(\cdot\mid q,h_t,\mathcal{K}_{\mathrm{rag}})$ for $t>T_{\mathrm{rag}}$, such that comparable historical situations, rather than parametric knowledge alone, inform the final answer.

\definecolor{headerpurple}{HTML}{D9D7E3}
\definecolor{modelgray}{HTML}{E5E5E5}
\definecolor{altgray}{HTML}{F0F0F0}
\begin{table*}[!t]
\centering
\caption{Performance across atomic procedures. The best method is shown in bold and the second best is underlined in italics.}
\vspace{-5pt}
\resizebox{\linewidth}{!}{%
\scriptsize
\begin{tabular}{l|c|cc|c|cc|c|cc|c|cc|c|cc}
\toprule
\rowcolor{headerpurple}
\multirow{2}{*}{\cellcolor{white}\textbf{Method}} & \multicolumn{3}{c|}{\textbf{Event Characterization}} & \multicolumn{3}{c|}{\textbf{Spatiotemporal Prediction}} & \multicolumn{3}{c|}{\textbf{Impact Assessment}} & \multicolumn{3}{c|}{\textbf{Responsive Decision-Making}} & \multicolumn{3}{c}{\textbf{Overall}} \\
\cmidrule(lr){2-4} \cmidrule(lr){5-7} \cmidrule(lr){8-10} \cmidrule(lr){11-13} \cmidrule(lr){14-16}
& \textit{Avg.} & \textit{CPR} & \textit{EPR} & \textit{Avg.} & \textit{CPR} & \textit{EPR} & \textit{Avg.} & \textit{CPR} & \textit{EPR} & \textit{Avg.} & \textit{CPR} & \textit{EPR} & \textit{Avg.} & \textit{CPR} & \textit{EPR} \\
\midrule
\rowcolor{modelgray}\multicolumn{16}{c}{\texttt{\textbf{Qwen3.7-Plus}}} \\
\textbf{Zephyrus} & 0.453 & 1.000 & 1.000 & 0.199 & 1.000 & 0.981 & 0.461 & 1.000 & 1.000 & 0.252 & 1.000 & 1.000 & 0.329 & 1.000 & 0.995 \\
\rowcolor{altgray}
\textbf{EWE} & 0.422 & 1.000 & 0.922 & 0.218 & 0.994 & 0.956 & 0.448 & 1.000 & 0.984 & 0.221 & 1.000 & 0.981 & 0.318 & 0.998 & 0.969 \\
\textbf{ClimAgent} & 0.438 & 1.000 & 0.797 & 0.224 & 0.800 & 0.969 & 0.449 & 1.000 & 0.984 & 0.268 & 1.000 & 0.994 & 0.335 & 0.944 & 0.962 \\
\rowcolor{altgray}
\textbf{ClimateAgent} & 0.456 & 1.000 & 0.797 & \underline{\textit{0.260}} & 1.000 & 0.838 & 0.461 & 1.000 & 0.927 & 0.269 & 1.000 & 0.994 & 0.351 & 1.000 & 0.906 \\
\textbf{HVR-Met} & 0.403 & 1.000 & 0.891 & 0.241 & 1.000 & 0.969 & \underline{\textit{0.479}} & 1.000 & 1.000 & 0.254 & 0.994 & 0.988 & 0.342 & 0.998 & 0.976 \\
\midrule
\rowcolor{altgray}
\textbf{\model-Base} & \underline{\textit{0.509}} & 1.000 & 1.000 & 0.226 & 1.000 & 1.000 & 0.266 & 0.995 & 0.984 & 0.253 & 1.000 & 1.000 & 0.278 & 0.998 & 0.995 \\
\textbf{\model-RAG} & 0.488 & 1.000 & 0.938 & 0.230 & 1.000 & 0.875 & \textbf{0.521} & 1.000 & 0.948 & \underline{\textit{0.537}} & 1.000 & 0.825 & \underline{\textit{0.441}} & 1.000 & 0.892 \\
\rowcolor{altgray}
\textbf{\model-Skill} & 0.478 & 1.000 & 0.953 & 0.225 & 0.994 & 0.944 & 0.476 & 1.000 & 0.964 & 0.472 & 0.994 & 0.912 & 0.406 & 0.997 & 0.943 \\
\textbf{\model-Modeling} & \textbf{0.522} & 1.000 & 0.984 & \textbf{0.277} & 1.000 & 1.000 & 0.463 & 1.000 & 0.984 & \textbf{0.554} & 1.000 & 0.994 & \textbf{0.443} & 1.000 & 0.991 \\
\midrule
\rowcolor{modelgray}\multicolumn{16}{c}{\texttt{\textbf{GPT-5.4 mini}}} \\
\textbf{Zephyrus} & \textbf{0.516} & 1.000 & 1.000 & 0.216 & 0.938 & 0.988 & 0.372 & 0.984 & 0.901 & 0.216 & 0.994 & 1.000 & 0.301 & 0.976 & 0.964 \\
\rowcolor{altgray}
\textbf{EWE} & 0.475 & 1.000 & 0.906 & \underline{\textit{0.248}} & 0.975 & 0.988 & 0.382 & 0.917 & 0.969 & 0.195 & 1.000 & 0.994 & 0.303 & 0.965 & 0.974 \\
\textbf{ClimAgent} & 0.481 & 1.000 & 0.641 & 0.245 & 0.994 & 0.925 & 0.401 & 1.000 & 0.953 & 0.240 & 1.000 & 0.975 & 0.322 & 0.998 & 0.917 \\
\rowcolor{altgray}
\textbf{ClimateAgent} & 0.484 & 0.969 & 0.812 & 0.185 & 0.887 & 0.875 & 0.392 & 0.974 & 0.896 & 0.216 & 0.994 & 0.919 & 0.296 & 0.955 & 0.887 \\
\textbf{HVR-Met} & 0.263 & 0.625 & 0.812 & 0.180 & 0.750 & 0.931 & 0.357 & 0.844 & 0.922 & 0.190 & 0.938 & 0.988 & 0.251 & 0.819 & 0.931 \\
\midrule
\rowcolor{altgray}
\textbf{\model-Base} & 0.512 & 1.000 & 1.000 & 0.227 & 1.000 & 1.000 & 0.391 & 1.000 & 0.990 & 0.224 & 1.000 & 1.000 & 0.313 & 1.000 & 0.997 \\
\textbf{\model-RAG} & \textbf{0.516} & 1.000 & 0.922 & \textbf{0.285} & 1.000 & 0.925 & \textbf{0.530} & 1.000 & 0.896 & \textbf{0.363} & 1.000 & 0.950 & \textbf{0.414} & 1.000 & 0.922 \\
\rowcolor{altgray}
\textbf{\model-Skill} & 0.391 & 1.000 & 0.953 & 0.247 & 1.000 & 0.925 & 0.427 & 1.000 & 0.953 & \underline{\textit{0.361}} & 1.000 & 0.919 & 0.355 & 1.000 & 0.936 \\
\textbf{\model-Modeling} & 0.472 & 1.000 & 1.000 & 0.238 & 1.000 & 0.994 & \underline{\textit{0.527}} & 1.000 & 1.000 & 0.331 & 1.000 & 1.000 & \underline{\textit{0.386}} & 1.000 & 0.998 \\
\midrule
\rowcolor{modelgray}\multicolumn{16}{c}{\texttt{\textbf{Gemini 3.1 Flash-Lite}}} \\
\textbf{Zephyrus} & 0.438 & 1.000 & 0.797 & 0.162 & 0.787 & 0.944 & 0.363 & 1.000 & 0.891 & 0.214 & 1.000 & 0.956 & 0.274 & 0.941 & 0.913 \\
\rowcolor{altgray}
\textbf{EWE} & 0.491 & 1.000 & 0.562 & 0.233 & 1.000 & 0.613 & 0.384 & 1.000 & 0.932 & 0.201 & 1.000 & 0.900 & 0.303 & 1.000 & 0.793 \\
\textbf{ClimAgent} & \underline{\textit{0.534}} & 1.000 & 0.688 & 0.233 & 1.000 & 0.600 & 0.375 & 1.000 & 0.870 & 0.209 & 1.000 & 0.994 & 0.307 & 1.000 & 0.809 \\
\rowcolor{altgray}
\textbf{ClimateAgent} & 0.512 & 0.984 & 0.453 & 0.209 & 1.000 & 0.512 & 0.397 & 1.000 & 0.807 & 0.191 & 1.000 & 0.938 & 0.300 & 0.998 & 0.722 \\
\textbf{HVR-Met} & 0.484 & 1.000 & 0.938 & 0.247 & 1.000 & 0.800 & 0.381 & 1.000 & 0.984 & 0.237 & 1.000 & 0.988 & 0.316 & 1.000 & 0.929 \\
\midrule
\rowcolor{altgray}
\textbf{\model-Base} & \textbf{0.544} & 1.000 & 1.000 & \underline{\textit{0.263}} & 1.000 & 1.000 & 0.363 & 1.000 & 1.000 & 0.224 & 1.000 & 1.000 & 0.317 & 1.000 & 1.000 \\
\textbf{\model-RAG} & \underline{\textit{0.534}} & 1.000 & 1.000 & \textbf{0.269} & 1.000 & 0.994 & \textbf{0.485} & 1.000 & 0.896 & \textbf{0.411} & 1.000 & 0.963 & \textbf{0.410} & 1.000 & 0.953 \\
\rowcolor{altgray}
\textbf{\model-Skill} & 0.506 & 1.000 & 1.000 & 0.248 & 1.000 & 1.000 & \underline{\textit{0.413}} & 1.000 & 0.990 & 0.247 & 1.000 & 1.000 & 0.332 & 1.000 & 0.997 \\
\textbf{\model-Modeling} & 0.484 & 1.000 & 0.969 & 0.241 & 1.000 & 0.981 & 0.407 & 1.000 & 1.000 & \underline{\textit{0.300}} & 1.000 & 0.975 & \underline{\textit{0.340}} & 1.000 & 0.984 \\
\bottomrule
\end{tabular}
}
\label{tab:atomic_aggregate_performance}
\end{table*}

\subsubsection{\model-Skill}
\model-Skill uses historical cases as practice tasks to acquire reusable solution procedures.
The harness consists of three phases: retrieval, rehearsal, and target solving.
First, a rehearsal-selection instruction $I_{\mathrm{reh}}$ guides the agent to retrieve $\mathcal{K}_{\mathrm{reh}}=\{\mathcal{K}_{j}\}_{j=1}^{m}\subset\mathcal{K}$, where the selected cases exercise reasoning patterns relevant to the current task.
Second, for each case $\mathcal{K}_j=(q_j,y_j)$, the agent solves $q_j$ without access to $y_j$, using a compact skill summary $\mathcal{S}_j$ that contains previously distilled textual guidance.
After rehearsal, the reference answer is revealed only to update the skills, yielding $\mathcal{S}_{j+1}=\pi_{\theta}(I_{\mathrm{skill}},\mathcal{S}_j,h^{j}_{T_j},y_j)$, where $I_{\mathrm{skill}}$ specifies how to compare the rehearsal trajectory with the reference answer and distill reusable procedural guidance for skill refinement.
Finally, the target task is solved with the accumulated skills under the policy $\pi_{\theta}(\cdot\mid q,h_t,\mathcal{S}_{m+1})$, allowing prior cases to improve the solution procedure rather than serve as direct answer examples.

\subsubsection{\model-Modeling}
\label{sec:modeling-agent}
\model-Modeling introduces a modeling harness that converts historical cases into task-specific predictive evidence.
Guided by a modeling instruction $I_{\mathrm{model}}$, the agent constructs training and validation samples from $\mathcal{K}$, selects an appropriate ML model using the enabled ML development tools, and follows $\{r_t,a_t\}\sim\pi_{\theta}(\cdot\mid q,h_t,I_{\mathrm{model}})$ until execution returns a trained predictor $M$ together with the observation at turn $T_{\mathrm{mod}}$, denoted by $\{o_{T_{\mathrm{mod}}},M\}=\mathcal{E}(a_{T_{\mathrm{mod}}})$.
The subsequent solving phase follows the base workflow with $M$ as an additional predictive signal, using $\{r_t,a_t\}\sim\pi_{\theta}(\cdot\mid q,h_t,M)$ for $t>T_{\mathrm{mod}}$.
This variant targets tasks in which recurring patterns across historical events provide useful statistical evidence for operational judgment.
\begin{figure*}[ht]
    \centering
    \includegraphics[width=0.85\linewidth]{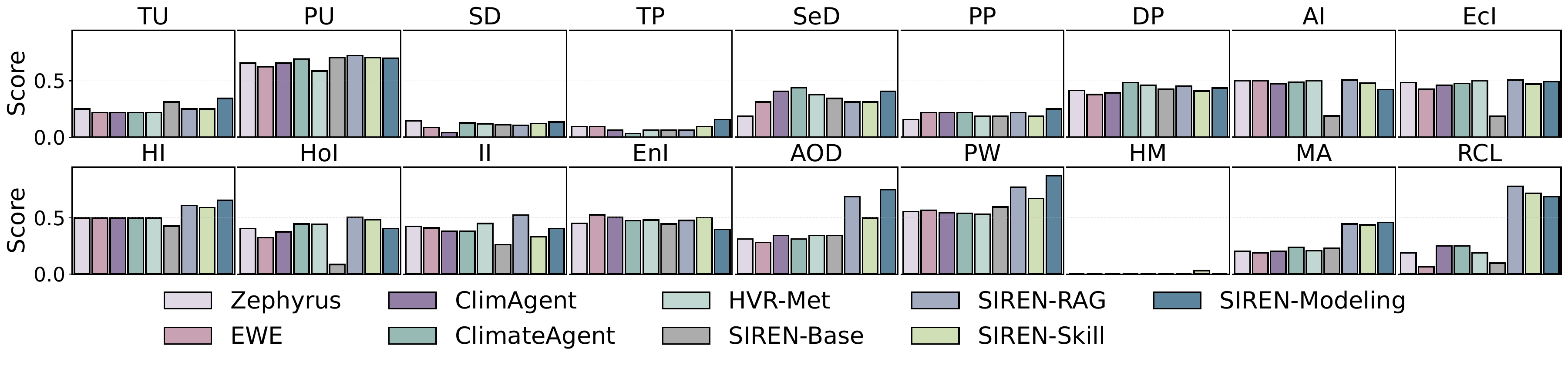}
    \vspace{-5pt}
    \caption{Subtask-level performance, with full subtask names and detailed descriptions provided in \appref{app:task_taxonomy}.}
    \label{figure:atomic_subtask}
\end{figure*}

\section{Experiments}


In this section, we evaluate the efficacy of \model\xspace agents for extreme-weather warning from the perspectives of individual procedures and the end-to-end warning chains.
Specifically, we compare \model\xspace agents with five strong weather-science baselines: Zephyrus~\cite{zephyrus2026}, EWE~\cite{jiang2025ewe}, ClimAgent~\cite{climagent2026}, ClimateAgent~\cite{kim2025climateagent}, and HVR-Met~\cite{hvrmet2026}. 
We instantiate all agents with three advanced API-based LLM backbones: Qwen3.7-Plus~\cite{qwen37plus2026}, GPT-5.4 mini~\cite{gpt54mini2026}, and Gemini 3.1 Flash-Lite~\cite{gemini31flashlite2026}.
Implementation details are provided in \appref{app:implementation_details}.

\subsection{Individual Procedure Evaluation}

We evaluate individual task procedures from five perspectives: aggregate performance across task categories, fine-grained behavior on individual subtasks, robustness across event types, geographic variation across states, and temporal variation across months.

\subsubsection{Aggregate Performance}

\tabref{tab:atomic_aggregate_performance} reports the aggregate results for the four individual task categories and three LLM backbones, yielding five main observations:
(1) \textit{The benchmark remains challenging, with difficulty varying across operational procedures.}
The best Overall Score is only 0.443, with most results below 0.4.
Event characterization is relatively tractable, whereas prediction is consistently the hardest category. Impact assessment and decision-making benefit more substantially from historical experience.
(2) \textit{Grounding agents in historical experience substantially improves their operational capabilities.}
The best \model\xspace variant outperforms the strongest baseline by 26.2\%, 28.6\%, and 29.7\% with Qwen, GPT, and Gemini, respectively.
Together with the consistent gains of \model-RAG, \model-Skill, and \model-Modeling over \model-Base, these results demonstrate the importance of historical experience for operational reasoning.
(3) \textit{The execution environment itself provides a strong foundation for weather-oriented agents.}
Without historical cases, \model-Base attains Overall Scores of 0.313 and 0.317 on GPT and Gemini: it is within 0.009 of the strongest GPT baseline and narrowly exceeds the strongest Gemini baseline.
Its lower Qwen score of 0.278 nevertheless indicates that the environment provides a competitive foundation but cannot replace the use of historical experience.
(4) \textit{The three experience-grounded mechanisms exhibit distinct and complementary strengths.}
\model-RAG achieves the best performance with GPT and Gemini, whereas \model-Modeling ranks first with Qwen and is particularly competitive in prediction and decision-making.
\model-Skill generally improves over \model-Base but underperforms the other variants, suggesting that reusable skills preserve less task-specific information than analogies or predictive models.
(5) \textit{Answer quality and execution reliability must be assessed separately.}
Across all backbones, the \model\xspace agents maintain CPR values from 0.997 to 1.000, while EPR values range from 0.892 to 1.000.
Thus, experience grounding rarely prevents task completion, but some high-scoring variants still incur code-execution failures that impose reliability costs.

\subsubsection{Subtask-Level Analysis}
\label{sec:atomic_subtask_analysis}

\figref{figure:atomic_subtask} presents a fine-grained comparison across the 18 atomic subtasks using Qwen3.7-Plus.
Historical experience benefits all four operational stages, although no mechanism dominates every subtask.
\model-RAG performs best on most impact-assessment tasks, whereas \model-Modeling performs particularly well on temporal and path prediction and several warning and resource-allocation tasks.
\model-Skill remains competitive on selected impact and mitigation tasks, while the consistently weak hazard-mitigation results reveal a shared difficulty in deriving interventions from limited context.
These differences further confirm the complementarity of the three experience mechanisms.

\subsubsection{Event-Type Analysis}

\begin{figure}[h]
    \centering
    \includegraphics[width=0.85\linewidth]{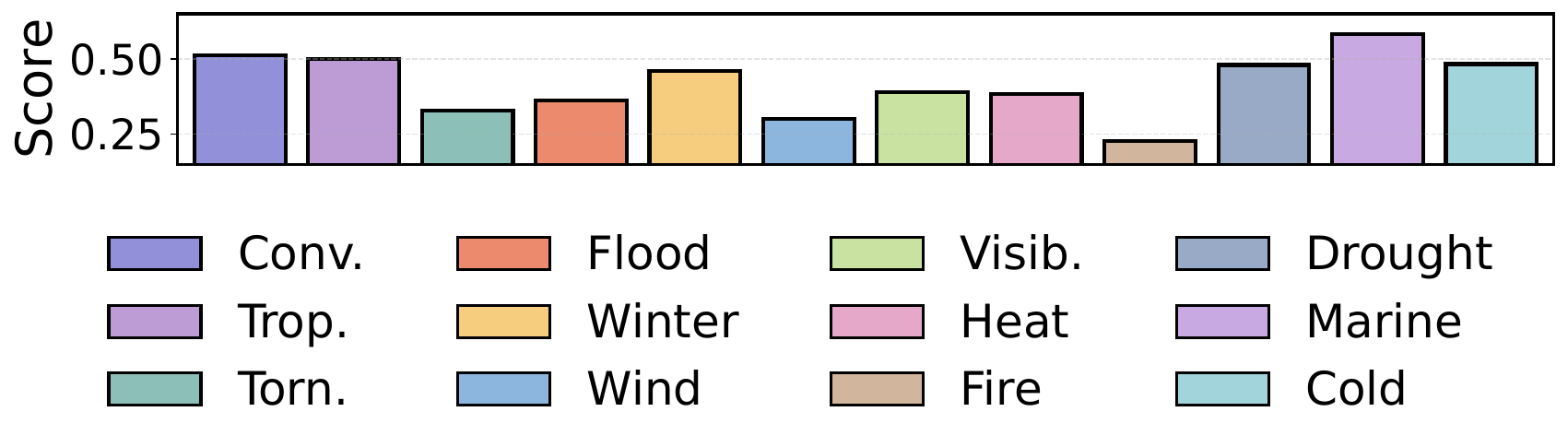}
    \vspace{-5pt}
    \caption{Performance across event types.}
    \vspace{-5pt}
\label{figure:atomic_event_type}
\end{figure}

\figref{figure:atomic_event_type} examines the aggregate performance of \model-RAG with Qwen3.7-Plus across extreme event families, with definitions of the 12 event families provided in \appref{app:extreme_weather_events}.
Experience grounding transfers across diverse hazards rather than benefiting only frequent event families.
Performance is strongest for marine hazards and remains competitive for convective storms, tropical systems, and several cold-season hazards.
Fire, high-wind events, and tornadoes are more challenging, likely because their rapid development or localized impacts complicate case retrieval and evidence transfer.
This variation highlights the need for stronger robustness across hazard-specific dynamics.
Complete method comparisons are provided in \appref{app:event_type_results}.

\subsubsection{State-Level Analysis Across Regions}

\begin{figure}[h]
    \centering
    \includegraphics[width=0.85\linewidth]{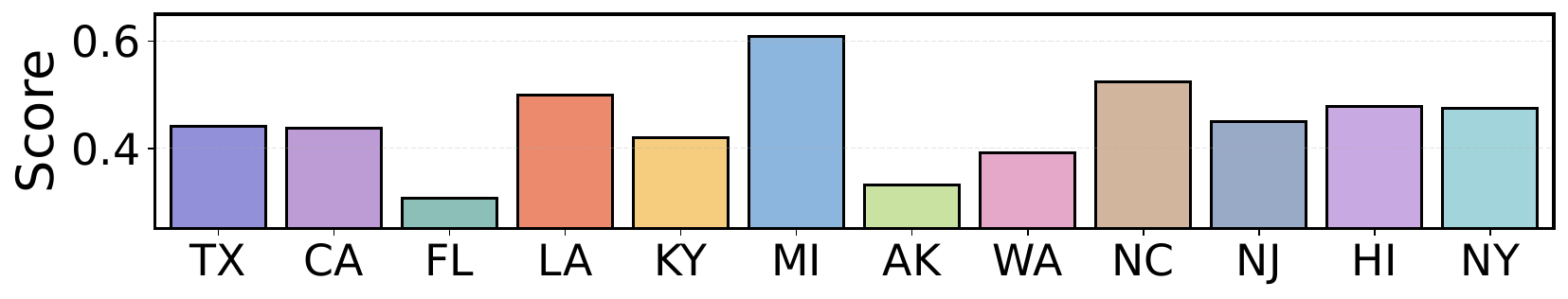}
    \vspace{-5pt}
    \caption{Performance across states.}
    \vspace{-5pt}
    \label{figure:atomic_region}
\end{figure}

\figref{figure:atomic_region} compares the performance of \model-RAG with Qwen3.7-Plus across twelve representative states.
Michigan achieves the best performance, followed by strong results in North Carolina and Louisiana, whereas Florida and Alaska are more challenging.
The pattern does not follow a simple regional divide: performance varies substantially within the Gulf Coast and Southeast, while New York and New Jersey remain competitive in the Northeast.
These differences likely reflect regional event compositions and the alignment of historical cases with local conditions, highlighting the need for stronger geographic robustness.
Complete state-level and joint event--state comparisons are provided in \appref{app:state_results} and \appref{app:event_state_results}.

\subsubsection{Temporal Analysis Across Months}

\begin{figure}[ht]
    \centering
    \includegraphics[width=0.8\linewidth]{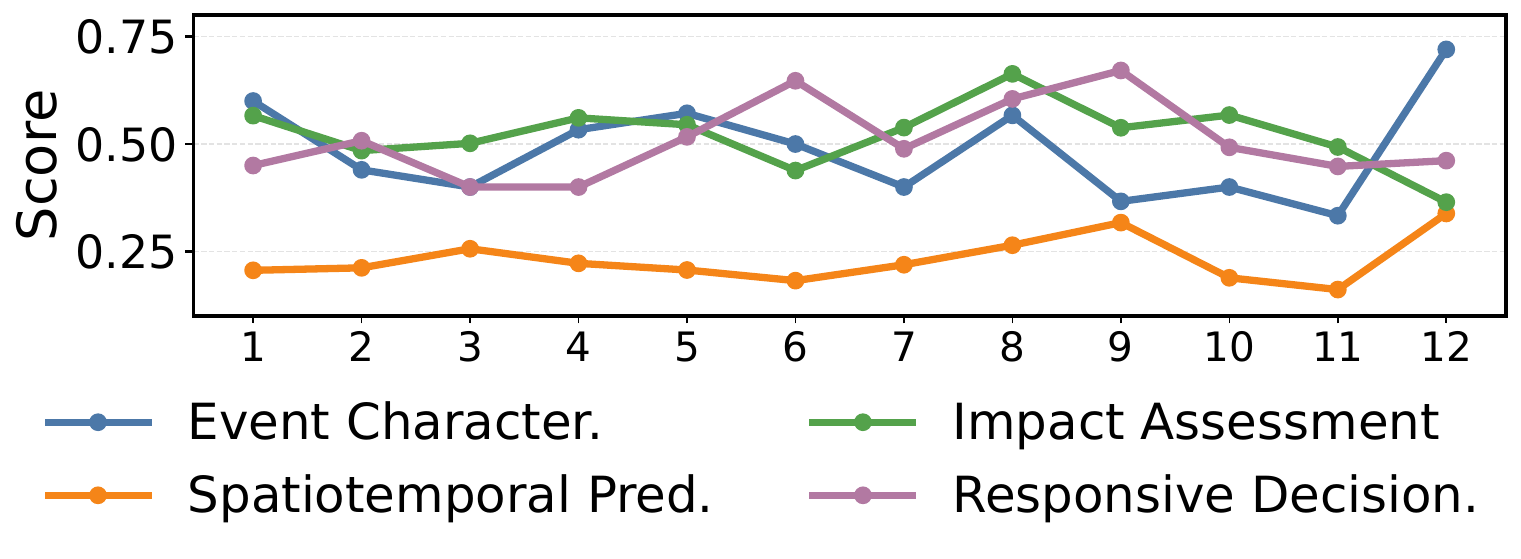}
    \vspace{-5pt}
    \caption{Performance across months.}
    \vspace{-5pt}
    \label{figure:atomic_temporal}
\end{figure}

\figref{figure:atomic_temporal} shows the monthly performance of \model-RAG with Qwen3.7-Plus in each task category.
Event characterization and impact assessment vary noticeably throughout the year, reflecting seasonal changes in event composition and the efficacy of analogous cases.
Decision-making performs better during the middle and later parts of the year, whereas prediction is more stable but consistently more difficult than the other categories.
Overall, experience grounding remains effective throughout the year, although the remaining variation highlights the need for stronger temporal robustness.
Complete monthly comparisons are provided in \appref{app:month_results}.

\subsection{End-to-end Chain Evaluation}

We further evaluate complete operational task chains that sequentially solve four individual procedures for the same event.

\subsubsection{End-to-end Performance}

\definecolor{headerpurple}{HTML}{D9D7E3}
\definecolor{modelgray}{HTML}{E5E5E5}
\definecolor{altgray}{HTML}{F0F0F0}
\begin{table}[!t]
\centering
\caption{Operational-chain performance with EC, SP, IA, and RD denoting four task procedures, respectively.}
\resizebox{\linewidth}{!}{%
\scriptsize
\begin{tabular}{l|c|c|c|c|c}
\toprule
\rowcolor{headerpurple}
\textbf{Method} & \textbf{EC} & \textbf{SP} & \textbf{IA} & \textbf{RD} & \textbf{Overall} \\
\midrule
\textbf{Zephyrus} & 0.317 & 0.083 & 0.025 & 0.536 & 0.240 \\
\rowcolor{altgray}
\textbf{EWE} & 0.225 & 0.083 & 0.180 & 0.586 & 0.268 \\
\textbf{ClimAgent} & 0.300 & \underline{\textit{0.125}} & 0.136 & 0.583 & 0.286 \\
\rowcolor{altgray}
\textbf{ClimateAgent} & 0.258 & \underline{\textit{0.125}} & 0.172 & 0.472 & 0.257 \\
\textbf{HVR-Met} & 0.275 & \underline{\textit{0.125}} & 0.124 & 0.491 & 0.254 \\
\midrule
\rowcolor{altgray}
\textbf{\model-Base} & \underline{\textit{0.325}} & 0.083 & 0.107 & 0.558 & 0.268 \\
\textbf{\model-RAG} & \textbf{0.342} & \textbf{0.167} & \underline{\textit{0.275}} & \textbf{0.731} & \textbf{0.379} \\
\rowcolor{altgray}
\textbf{\model-Skill} & 0.233 & \underline{\textit{0.125}} & 0.252 & \underline{\textit{0.602}} & \underline{\textit{0.303}} \\
\textbf{\model-Modeling} & 0.258 & \underline{\textit{0.125}} & \textbf{0.305} & 0.332 & 0.255 \\
\bottomrule
\end{tabular}
}
\label{tab:chain_aggregate_oldest_baselines_gemini_3_1_flash_lite_score}
\end{table}

\tabref{tab:chain_aggregate_oldest_baselines_gemini_3_1_flash_lite_score} reports chain-level performance with Gemini 3.1 Flash-Lite.
Chain-level performance is generally lower than isolated-procedure performance, confirming the greater difficulty of end-to-end warning chains.
\model-RAG ranks first overall with a score of 0.379, outperforming the strongest baseline, ClimAgent, by 32.5\%. 
\model-Skill ranks second with a score of 0.303.
\model-RAG leads EC, SP, and RD, whereas \model-Modeling leads IA, demonstrating the complementary benefits of retrieved cases and learned predictive evidence.
The weaker overall results of \model-Modeling indicate that historical experience must be used effectively rather than merely made available.

\subsubsection{Cross-Procedure Dependency}

\tabref{tab:chain_dependency_spearman_oldest_baselines_gemini_3_1_flash_lite_siren_rag} reports inter-stage dependencies for \model-RAG with Gemini 3.1 Flash-Lite.
Spearman $\rho$ measures the association between each procedure, or its cumulative prefix, and the subsequent procedure across all operational chains.
All five dependencies are positive, with the strongest association observed for $EC/SP\!\rightarrow\!IA$ at 0.589. 
The positive associations for $IA\!\rightarrow\!RD$ and $EC/SP/IA\!\rightarrow\!RD$ further indicate that stronger upstream performance tends to improve the final decision.
These consistent dependencies confirm that performance propagates through the end-to-end warning chain, highlighting the importance of complete operational chains beyond isolated tasks.

\begin{table}[!t]
\centering
\caption{Cross-procedure dependency.}
\resizebox{0.6\linewidth}{!}{%
\scriptsize
\begin{tabular}{lc}
\toprule
\rowcolor{headerpurple}
\textbf{Dependency} & \textbf{Spearman $\rho$} \\
\midrule
$EC\!\rightarrow\!SP$ & 0.196 \\
\rowcolor{altgray}
$SP\!\rightarrow\!IA$ & 0.226 \\
$IA\!\rightarrow\!RD$ & 0.282 \\
\rowcolor{altgray}
$(EC/SP)\!\rightarrow\!IA$ & 0.589 \\
$(EC/SP/IA)\!\rightarrow\!RD$ & 0.254 \\
\bottomrule
\end{tabular}
}
\label{tab:chain_dependency_spearman_oldest_baselines_gemini_3_1_flash_lite_siren_rag}
\end{table}

\section{Related Work}
In this section, we review recent studies on benchmarks and LLM agents in weather science.

\noindent\textbf{Benchmarks for Weather Science.}
Early benchmarks focus on numerical forecasting and event detection~\cite{weatherbench2024,hrextreme2025,extremeweather2017}, supporting meteorological prediction and hazard localization~\cite{personal2024,scaling2025,uniextreme2026}.
LLM-oriented datasets extend evaluation to QA tasks, including weather reasoning~\cite{ma2024weatherqa,kmetbench2026}, anomaly interpretation~\cite{climateiqa2025}, event forecasting~\cite{cllmate2025}, warning-message generation~\cite{crisitext2026}, and forecast reporting~\cite{weathersyn2026}, but they typically provide the required inputs rather than requiring active evidence collection, tool use, and multi-stage reasoning.
Agent benchmarks more closely reflect realistic workflows: ZephyrusBench~\cite{zephyrus2026} covers diverse weather-science problems, ClimaBench \cite{climagent2026} focuses on research-oriented climate analysis, and Climate-Agent-Bench-85~\cite{kim2025climateagent} evaluates executable workflows and report generation.
Extreme-weather benchmarks further assess signal identification, physical explanation, and structured diagnosis~\cite{jiang2025ewe,hvrmet2026}.
Nevertheless, these benchmarks remain focused on scientific analysis rather than operational weather applications, \eg extreme-weather early warning, which require impact-oriented assessment and warning decisions.

\noindent\textbf{LLM Agents for Weather Science.}
Recent LLM agents automate weather and climate workflows through tool use, code execution, and domain-specific data access.
Zephyrus~\cite{zephyrus2026} provides unified Python interfaces to meteorological data, forecasting models, and simulations, while ClimAgent~\cite{climagent2026} supports planning and refinement for open-ended climate analysis.
ClimateAgent~\cite{kim2025climateagent} coordinates specialized agents to decompose climate questions and self-correct the resulting code, visualizations, and reports.
For extreme-weather diagnosis, EWE~\cite{jiang2025ewe} combines knowledge-guided planning with meteorological tools, whereas HVR-Met~\cite{hvrmet2026} iteratively verifies anomalous signals and replans its reasoning.
\section{Conclusion}
In this paper, we formulate extreme-weather early warning as an agentic problem and construct \bench, a benchmark that covers four atomic warning procedures and an end-to-end warning chain.
To address this problem, we develop \model, an experience-grounded agent framework that combines an execution environment that integrates heterogeneous weather evidence and analytical tools with a family of agents that leverage historical cases through retrieval, skill distillation, and predictive modeling.
Experiments with multiple LLM backbones and strong weather-agent baselines demonstrate the effectiveness of \model\xspace on both atomic tasks and operational warning chains.

\section*{Limitations and Ethical Considerations}

\noindent\textbf{Limitations.}
Our study is limited by (1) its focus on U.S. extreme-weather events and selected data sources, which may restrict generalization across regions, periods, and hazard regimes; and (2) the gap between retrospective benchmark evaluation and real-world deployment, where data latency, missing data, evolving forecasts, and operational constraints may affect performance.

\noindent\textbf{Ethical considerations.}
Our study (1) uses only publicly available governmental and geospatial data in accordance with the stated access and attribution requirements, without identifiable private information or direct human-subject participation, and therefore did not require IRB review under our institutional policy; and (2) may be misused or produce incorrect warnings, public alarm, or resource misallocation, so SIREN is intended only to support qualified professionals who retain human oversight and decision authority.

\section*{Generative AI Usage}

Apart from the LLMs used as research components described in this paper, generative AI tools assisted with code generation, language polishing, and the creation of figure icons unrelated to the reported results. All generated content was reviewed by the authors, who independently verified the experimental results and conclusions and take full responsibility for the manuscript.


\bibliographystyle{ACM-Reference-Format}
\bibliography{main}

\appendix

\section{Task Taxonomy}
\label{app:task_taxonomy}
\tabref{tab:task_taxonomy} summarizes the 18 atomic subtasks, their abbreviations, evaluation protocols, and representative question templates.

\begin{table*}[t]
\centering
\caption{Atomic task taxonomy in \bench. MC, NR, GL, and OG denote multiple-choice classification, numeric regression, geospatial localization, and open-ended generation, respectively.}
\label{tab:task_taxonomy}
\small
\setlength{\tabcolsep}{3.2pt}
\renewcommand{\arraystretch}{1.08}
\begin{tabular}{c|cccl}
\toprule
\textbf{Task Procedure} & \textbf{Subtask} & \textbf{Abbr.} & \textbf{Eval.} & \textbf{Representative Question} \\
\midrule
\multirow{2}{*}{\makecell[c]{Event\\Characterization}}
& Type Understanding & TU & MC & Which extreme-weather event type best describes this record? \\
\cline{2-5}
& Physical Understanding & PU & OG & What physical processes and forcing ingredients drive this event? \\
\midrule
\multirow{5}{*}{\makecell[c]{Spatiotemporal\\Prediction}}
& Spatial Detection & SD & GL & Which state and county are currently experiencing this event? \\
\cline{2-5}
& Temporal Prediction & TP & MC & How many hours from the reference time until the event starts? \\
\cline{2-5}
& Severity Detection & SeD & MC & What is the current severity level of this event? \\
\cline{2-5}
& Path Prediction & PP & MC & In which direction is this event expected to move? \\
\cline{2-5}
& Duration Prediction & DP & NR & How long will this event last, in hours? \\
\midrule
\multirow{6}{*}{\makecell[c]{Impact\\Assessment}}
& Agricultural Impact & AI & NR & How much crop damage will this event cause, in USD? \\
\cline{2-5}
& Economic Impact & EcI & NR & How much property damage will this event cause, in USD? \\
\cline{2-5}
& Human Impact & HI & NR & How many injuries will this event cause? \\
\cline{2-5}
& Household Impact & HoI & NR & How many households require FEMA housing assistance? \\
\cline{2-5}
& Infrastructure Impact & II & NR & How much public-assistance funding is needed, in USD? \\
\cline{2-5}
& Energy Impact & EnI & NR & How long will the county-level power outage last, in hours? \\
\midrule
\multirow{5}{*}{\makecell[c]{Responsive\\Decision-Making}}
& Alert Operation Decision & AOD & MC & Which operation code should be applied to the current alert? \\
\cline{2-5}
& Public Warning & PW & OG & What public response instruction should be issued for this event? \\
\cline{2-5}
& Hazard Mitigation & HM & MC & Which hazard-mitigation projects should be selected? \\
\cline{2-5}
& Mission Assignment & MA & OG & What mission-assignment response is needed for this event? \\
\cline{2-5}
& Recovery Center Location & RCL & GL & Where should disaster-recovery centers be activated? \\
\bottomrule
\end{tabular}%
\end{table*}

\section{Extreme Weather Events}
\label{app:extreme_weather_events}
\tabref{tab:extreme_weather_events} defines the 12 event families covered by \bench.
\begin{table*}[t]
\centering
\caption{Extreme-weather event families covered by \bench.}
\label{tab:extreme_weather_events}
\small
\setlength{\tabcolsep}{5pt}
\renewcommand{\arraystretch}{1.08}
\begin{tabularx}{0.92\textwidth}{>{\raggedright\arraybackslash}p{3.1cm}>{\centering\arraybackslash}p{1.0cm}>{\raggedright\arraybackslash}X}
\toprule
\textbf{Event Type} & \textbf{Abbr.} & \textbf{Description} \\
\midrule
Convective Storms & Conv. & Thunderstorms producing hazards such as hail, lightning, heavy rain, or damaging convective winds. \\
\cline{1-3}
Tropical Systems & Trop. & Hurricanes, tropical storms, tropical depressions, and related wind, rain, and surge hazards. \\
\cline{1-3}
Tornadoes & Torn. & Violently rotating columns of air extending from a convective cloud to the ground. \\
\cline{1-3}
Floods & Flood & Riverine, flash, coastal, or other inundation of normally dry areas. \\
\cline{1-3}
Winter Weather & Winter & Snow, blizzard, sleet, ice, freezing rain, and other hazardous winter precipitation. \\
\cline{1-3}
High-Wind Events & Wind & Damaging non-convective winds, strong-wind episodes, and wind advisories. \\
\cline{1-3}
Visibility Hazards & Visib. & Fog, dust, smoke, or related conditions that substantially reduce visibility. \\
\cline{1-3}
Heat & Heat & Excessive-heat events and heat waves posing elevated thermal risk. \\
\cline{1-3}
Fire & Fire & Wildfires and fire-weather conditions that support ignition or rapid spread. \\
\cline{1-3}
Drought & Drought & Prolonged precipitation or moisture deficits that cause water stress. \\
\cline{1-3}
Marine Hazards & Marine & Hazardous marine winds, waves, thunderstorms, or other conditions over coastal waters. \\
\cline{1-3}
Cold-Weather Events & Cold & Extreme cold, wind chill, freeze, and related low-temperature hazards. \\
\bottomrule
\end{tabularx}
\end{table*}

\section{Analytical Tools}
\label{app:analytical_tools}
\tabref{tab:analytical_tools} lists the tools available in our agentic environment. Access to historical cases is restricted to experience-grounded variants, while ML tools are restricted to \model-Modeling.
Notably, our agents use task-specific ML models (\eg weather forecasting, decision-tree, and regression models) as intermediate reasoning tools rather than standalone predictors, while its ultimate goal is to solve practical and operational warning tasks that are not directly comparable to the isolated prediction objectives of conventional ML baselines.

\begin{table*}[t]
\centering
\caption{Enabled analytical tools in the agentic environment.}
\label{tab:analytical_tools}
\small
\setlength{\tabcolsep}{3.2pt}
\renewcommand{\arraystretch}{1.08}
\resizebox{\linewidth}{!}{%
\begin{tabular}{c|ll}
\toprule
\textbf{Category} & \textbf{Tool} & \textbf{Description} \\
\midrule
\multirow{6}{*}{\makecell[l]{Evidence\\Indexing}}
& \texttt{access\_historical\_cases} & Retrieves percentile slices of prior-year cases for experience-grounded variants. \\
\cline{2-3}
& \texttt{access\_hrrr\_reanalysis} & Loads cached or newly fetched HRRR analysis fields as a reusable weather-data bundle. \\
\cline{2-3}
& \texttt{access\_osm\_data} & Retrieves indexed OpenStreetMap features or road networks for a target U.S. region. \\
\cline{2-3}
& \texttt{access\_run\_artifact} & Reads a previously saved artifact through the controlled run-data boundary. \\
\cline{2-3}
& \texttt{access\_spc\_mesoscale\_images} & Fetches SPC mesoscale-analysis imagery for a requested valid time and product set. \\
\cline{2-3}
& \texttt{save\_run\_artifact} & Persists structured data, text, tables, or arrays for reuse within the current run. \\
\midrule
\multirow{7}{*}{\makecell[l]{Meteorological\\Analysis}}
& \texttt{apply\_threshold\_analysis} & Creates a threshold mask and reports the fraction of a field meeting a specified condition. \\
\cline{2-3}
& \texttt{compute\_frontogenesis} & Computes a two-dimensional frontogenesis diagnostic from temperature and wind fields. \\
\cline{2-3}
& \texttt{compute\_layer\_profile\_metrics} & Derives layer diagnostics such as bulk shear, lapse rate, helicity, and relative humidity. \\
\cline{2-3}
& \texttt{compute\_wind\_diagnostics} & Computes wind speed, vorticity, and divergence from gridded wind components. \\
\cline{2-3}
& \texttt{describe\_meteorological\_analysis\_options} & Lists the supported meteorological diagnostics and threshold-analysis modes. \\
\cline{2-3}
& \texttt{inspect\_gridded\_data\_bundle} & Inspects a weather or diagnostic bundle and reports its available variables. \\
\cline{2-3}
& \texttt{summarize\_gridded\_field} & Produces compact descriptive statistics for a gridded field or masked region. \\
\midrule
\makecell[l]{Atmospheric\\Forecasting}
& \texttt{access\_model\_forecast} & Loads model forecast fields at a requested initialization time and lead time. \\
\midrule
\multirow{5}{*}{\makecell[l]{Visual\\Processing}}
& \texttt{annotate\_image} & Adds points, boxes, and text labels to an image artifact. \\
\cline{2-3}
& \texttt{crop\_image} & Extracts a rectangular image region using normalized coordinates. \\
\cline{2-3}
& \texttt{visualize\_field\_comparison} & Renders two compatible fields side by side or as their difference. \\
\cline{2-3}
& \texttt{visualize\_gridded\_field} & Renders a gridded weather or diagnostic field as a map image. \\
\cline{2-3}
& \texttt{zoom\_in\_image} & Crops and enlarges a selected image region for closer inspection. \\
\midrule
\multirow{3}{*}{\makecell[l]{Impact\\Modeling}}
& \texttt{estimate\_integrated\_impact} & Combines hazard, exposure, and vulnerability inputs into an impact proxy. \\
\cline{2-3}
& \texttt{rank\_impacted\_regions} & Orders regions by a selected impact-score field. \\
\cline{2-3}
& \texttt{summarize\_impact\_distribution} & Summarizes impact values with descriptive statistics and percentiles. \\
\midrule
\multirow{5}{*}{\makecell[l]{Geospatial\\Normalization}}
& \texttt{align\_bbox\_to\_weather\_grid} & Snaps a geographic bounding box to the enclosing HRRR grid cells. \\
\cline{2-3}
& \texttt{build\_region\_grid\_mask} & Rasterizes a U.S. administrative region onto the HRRR grid. \\
\cline{2-3}
& \texttt{list\_supported\_us\_geospatial\_layers} & Lists the prepared U.S. administrative layers available for spatial normalization. \\
\cline{2-3}
& \texttt{lookup\_point\_location} & Resolves coordinates to prepared U.S. administrative regions. \\
\cline{2-3}
& \texttt{normalize\_location\_reference} & Matches a U.S. place reference to a canonical administrative feature. \\
\midrule
\multirow{4}{*}{\makecell[l]{ML\\Development}}
& \texttt{describe\_ml\_development\_options} & Lists supported ML problem types, input forms, and model families. \\
\cline{2-3}
& \texttt{fit\_vulnerability\_curve} & Fits a vulnerability function relating hazard intensity to observed impact. \\
\cline{2-3}
& \texttt{predict\_with\_saved\_model} & Applies a trained model to inference data and optionally saves its predictions. \\
\cline{2-3}
& \texttt{train\_ml\_model} & Trains and evaluates a lightweight predictive model on agent-constructed data. \\
\bottomrule
\end{tabular}
}
\end{table*}

\section{Implementation Details}
\label{app:implementation_details}
For each backbone, all compared methods use deterministic decoding with temperature 0. We use identical benchmark inputs and the same evaluation pipeline for our methods and the reproduced baselines. Given the scale of the benchmark and the number of evaluated methods and backbone models, we run each configuration once and report the resulting performance.

All experience-grounded variants use the same pool of historical cases. \model-RAG retrieves up to six cases for the target task, whereas \model-Skill may select and execute up to three practice cases before producing an answer. Transient model-service failures, unsuccessful code or tool executions, and malformed action or answer formats trigger bounded retries with explicit error feedback. A run is considered unsuccessful if the corresponding retry budget is exhausted.

We use Qwen3.6-27B~\cite{qwen3627b2026} with temperature 0 for all LLM-based evaluation components and apply the same evaluator to every method. The core prompts are provided in \appref{app:prompts}.

\section{Comparative Results across Event Types}
\label{app:event_type_results}
\begin{figure*}[t]
    \centering
    \subfloat[Qwen3.7-Plus]{\includegraphics[width=0.9\linewidth]{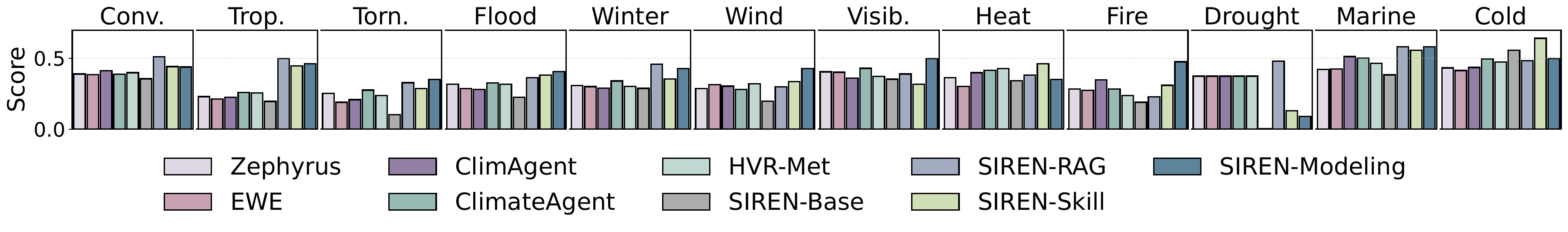}}\\
    \subfloat[GPT-5.4 mini]{\includegraphics[width=0.9\linewidth]{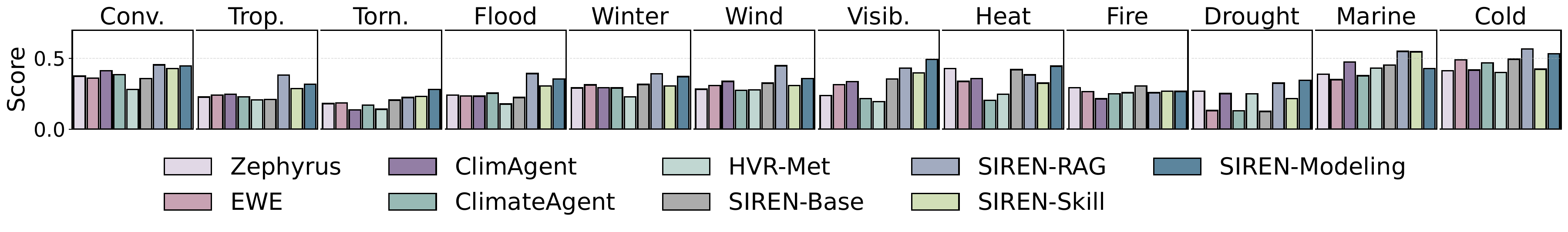}}\\
    \subfloat[Gemini 3.1 Flash-Lite]{\includegraphics[width=0.9\linewidth]{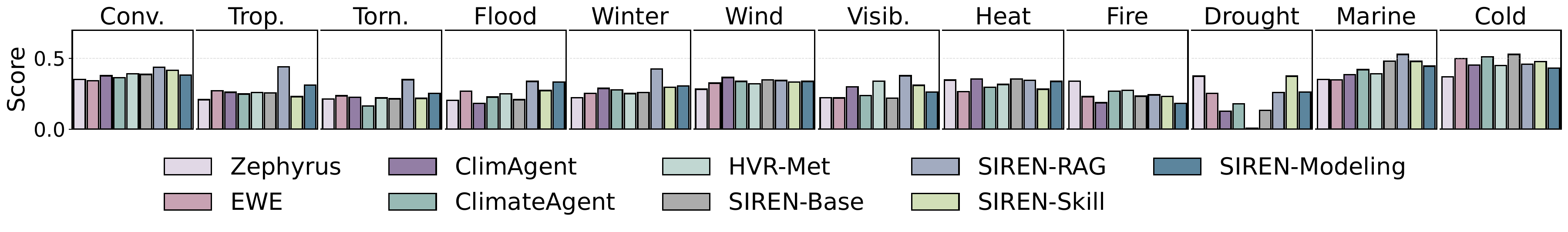}}
    \caption{Method comparison across event types. Each subfigure uses one LLM backbone and each panel represents one event family.}
    \label{fig:app_event_type_comparison}
\end{figure*}

\figref{fig:app_event_type_comparison} extends the aggregate event analysis to all methods and backbones. When averaged across the 12 event families, \model-Modeling performs marginally best with Qwen, whereas \model-RAG leads clearly with GPT and Gemini. The ordering varies across individual hazards, indicating that retrieved analogues and learned predictive evidence provide complementary benefits rather than uniform, event-independent gains.

\section{Comparative Results across States}
\label{app:state_results}
\begin{figure*}[t]
    \centering
    \subfloat[Qwen3.7-Plus]{\includegraphics[width=0.9\linewidth]{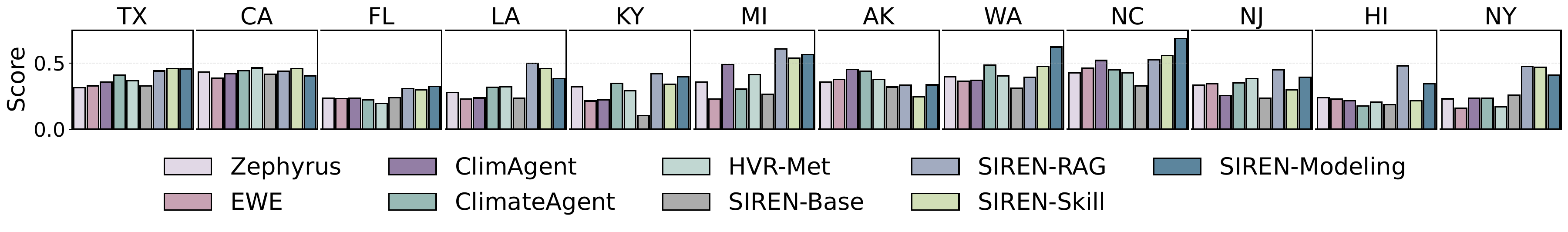}}\\
    \subfloat[GPT-5.4 mini]{\includegraphics[width=0.9\linewidth]{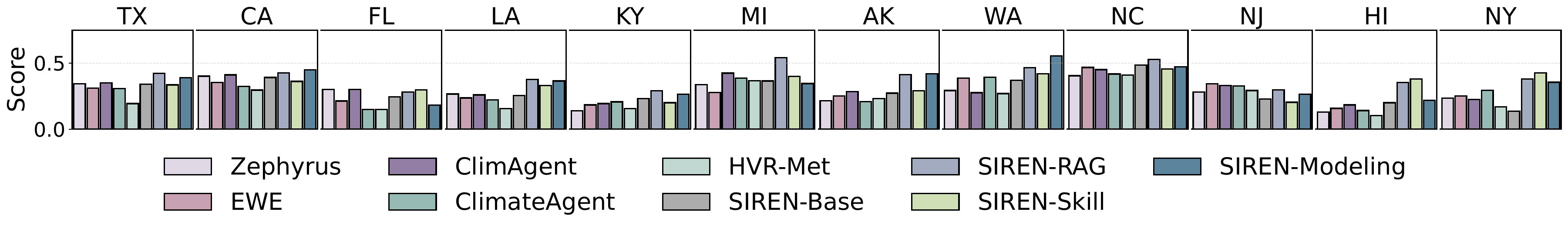}}\\
    \subfloat[Gemini 3.1 Flash-Lite]{\includegraphics[width=0.9\linewidth]{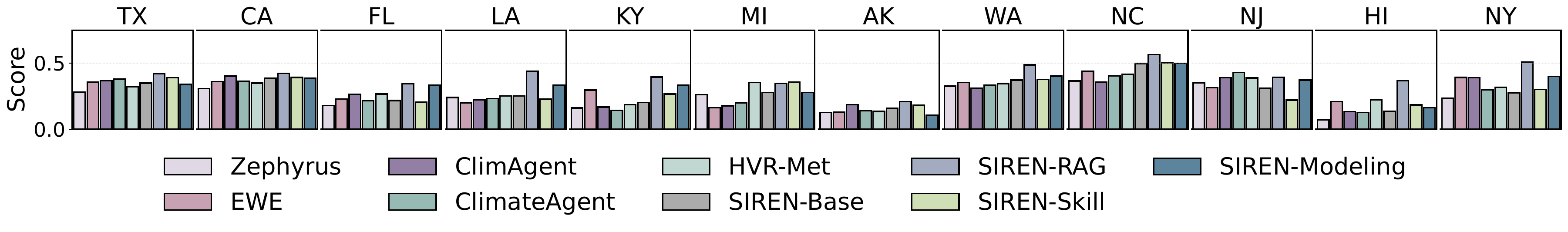}}
    \caption{Method comparison across the 12 most represented states.}
    \label{fig:app_state_comparison}
\end{figure*}

\figref{fig:app_state_comparison} shows that the benefits of historical experience extend broadly across geographic regions. \model-RAG achieves the highest mean across the selected states for all three backbones, narrowly outperforming \model-Modeling with Qwen and leading more clearly with GPT and Gemini.
The remaining variation across states indicates that agent performance is not yet geographically robust across heterogeneous local conditions.

\section{Comparative Results across Event Types and States}
\label{app:event_state_results}

We report the results for Qwen3.7-Plus on the four representative event families with the largest sample counts among those spanning at least 12 states. For each family, all 12 most represented states contain nonzero samples.

\begin{figure*}[t]
    \centering
    \includegraphics[width=0.9\linewidth]{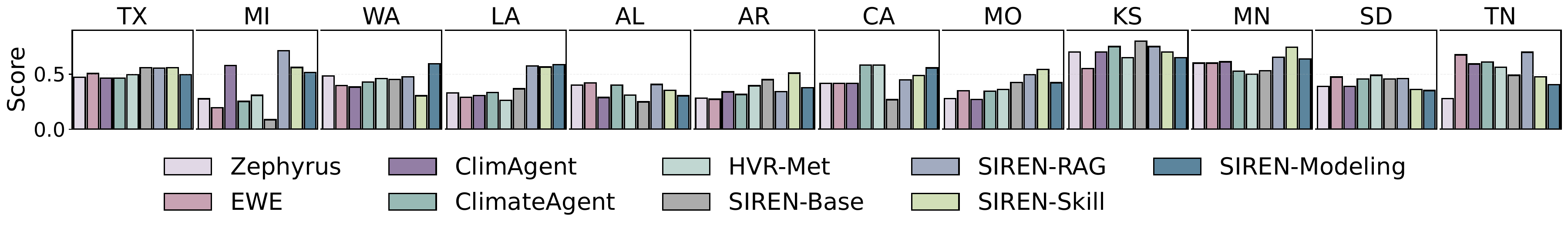}
    \caption{State-level method comparison for convective storms with Qwen3.7-Plus.}
    \label{fig:app_event_state_convective}
\end{figure*}

\begin{figure*}[t]
    \centering
    \includegraphics[width=0.9\linewidth]{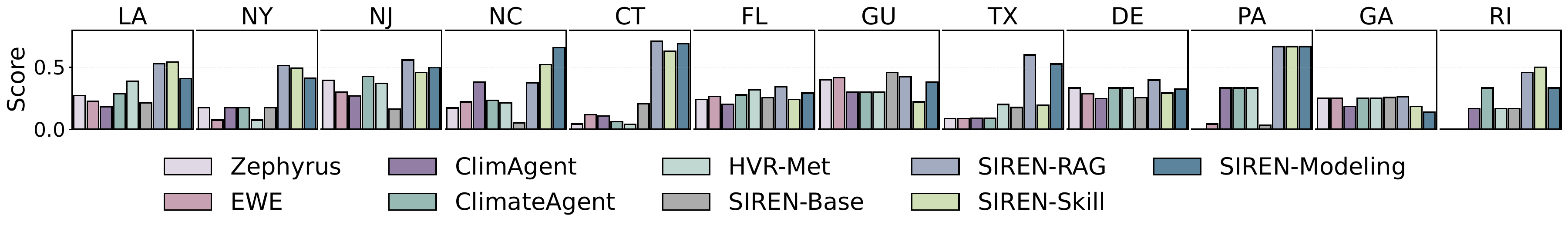}
    \caption{State-level method comparison for tropical systems with Qwen3.7-Plus.}
    \label{fig:app_event_state_tropical}
\end{figure*}

\begin{figure*}[t]
    \centering
    \includegraphics[width=0.9\linewidth]{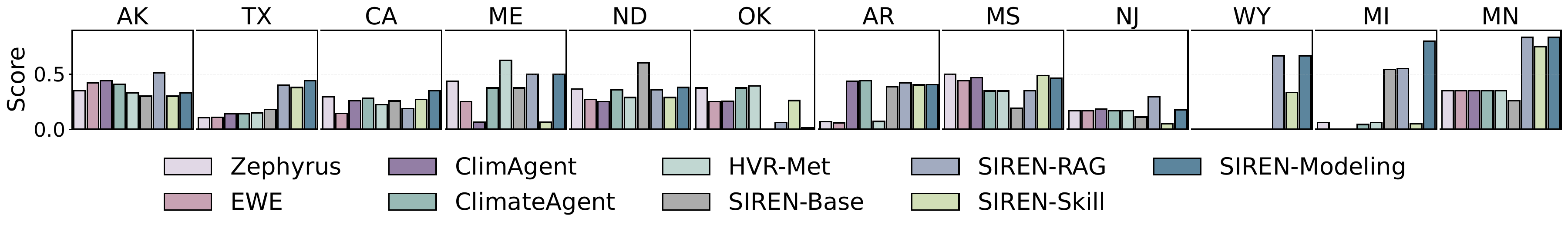}
    \caption{State-level method comparison for winter weather with Qwen3.7-Plus.}
    \label{fig:app_event_state_winter}
\end{figure*}

\begin{figure*}[t]
    \centering
    \includegraphics[width=0.9\linewidth]{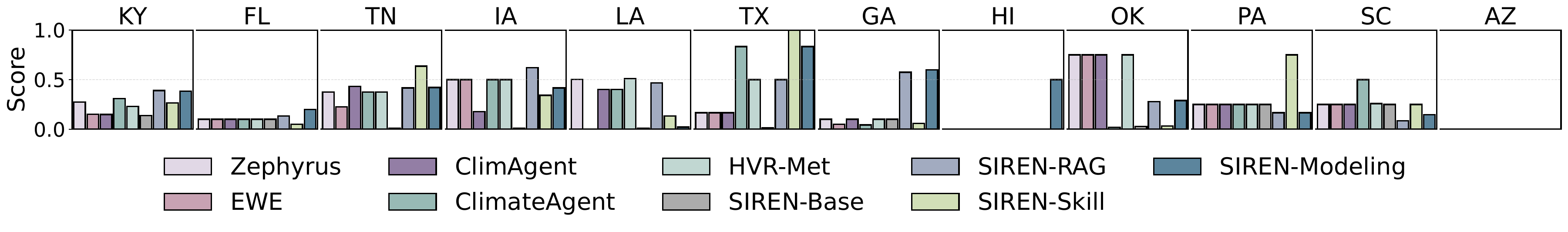}
    \caption{State-level method comparison for tornadoes with Qwen3.7-Plus.}
    \label{fig:app_event_state_tornado}
\end{figure*}

\figref{fig:app_event_state_convective}--\figref{fig:app_event_state_tornado} reveal substantial geographic variation within each event family. On average, \model-RAG performs best for convective storms and tropical systems, whereas \model-Modeling performs best for winter weather and tornadoes. 
The varying margins across states indicate that agent robustness depends jointly on event-specific dynamics and heterogeneous local conditions.

\section{Comparative Results across Months}
\label{app:month_results}
\begin{figure*}[t]
    \centering
    \subfloat[Qwen3.7-Plus]{\includegraphics[width=0.9\linewidth]{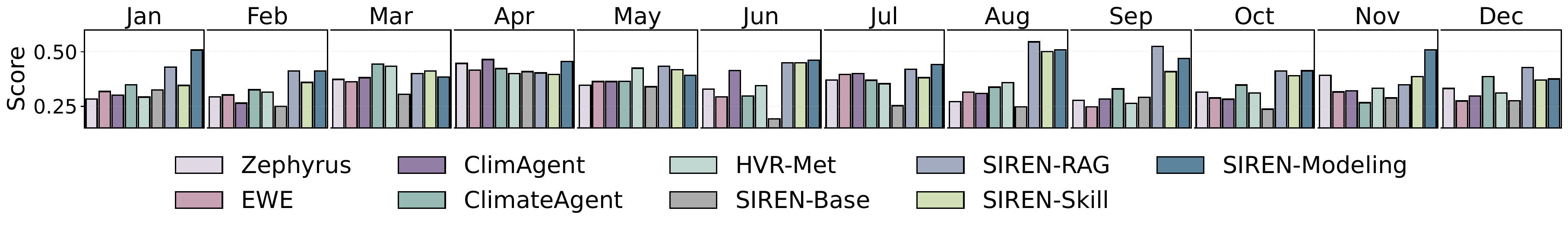}}\\
    \subfloat[GPT-5.4 mini]{\includegraphics[width=0.9\linewidth]{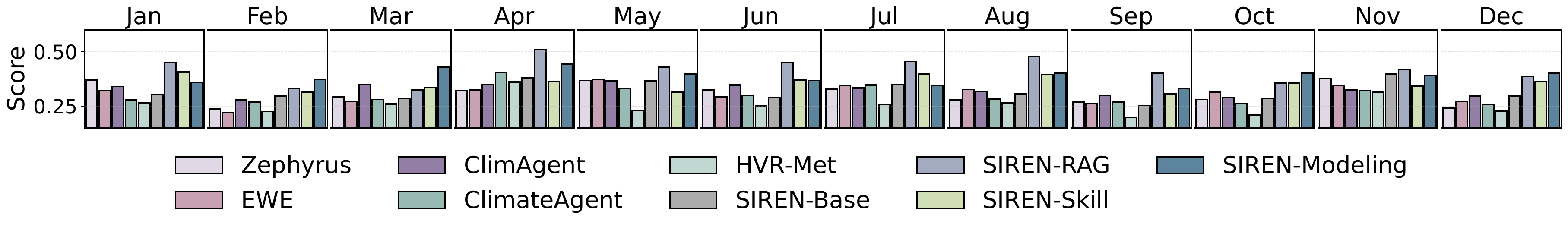}}\\
    \subfloat[Gemini 3.1 Flash-Lite]{\includegraphics[width=0.9\linewidth]{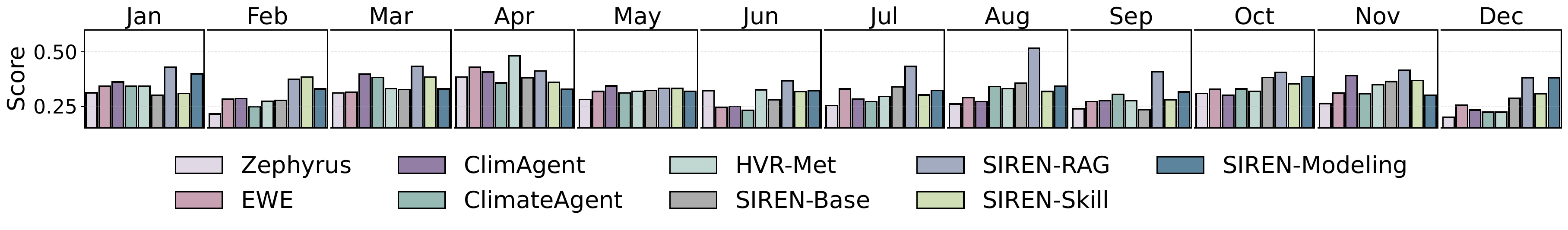}}
    \caption{Method comparison across calendar months.}
    \label{fig:app_month_comparison}
\end{figure*}

\figref{fig:app_month_comparison} demonstrates that experience grounding remains effective throughout the annual cycle. \model-Modeling achieves the highest monthly performance with Qwen, whereas \model-RAG leads with GPT and Gemini.
No baseline consistently dominates across all months, and the varying margins among the three SIREN mechanisms indicate uneven robustness to seasonal changes in event composition.

\section{Human Evaluation}
\label{app:human_evaluation}

We validate both LLM-based evaluators using \model-RAG outputs for all queries in each relevant open-ended setting and across all three backbone models. The recall-based study covers Public Warning and Mission Assignment, whereas the 0--10 study covers Physical Understanding. For each response, annotators assess the analysis and score produced by the evaluator, independently assign a human score under the same guidelines, and reach a binary consensus judgment. We measure LLM--human alignment using (1) \textbf{Agreement Rate}, the percentage of evaluator judgments considered reasonable by the annotators, (2) \textbf{Kendall Tau-b Rank Correlation Coefficient}, which measures agreement in pairwise ordering while accounting for tied scores, and (3) \textbf{Spearman Rank Correlation Coefficient}, which measures the monotonic association between the complete evaluator and human score rankings. \tabref{tab:human_eval_recall} and \tabref{tab:human_eval_scale} report the alignment results, indicating that both evaluators generally align with human rankings.

\begin{table}[ht]
\centering
\caption{LLM--human alignment performance for the recall-based evaluator (\%).}
\label{tab:human_eval_recall}
\setlength{\tabcolsep}{5pt}
\begin{tabular}{ccc}
\toprule
\textbf{Agree. Rate} & \textbf{Kendall Tau} & \textbf{Spearman Coeff.} \\
\midrule
73.96 & 80.50 & 87.05 \\
\bottomrule
\end{tabular}
\end{table}

\begin{table}[ht]
\centering
\caption{LLM--human alignment performance for the 0--10 scale evaluator (\%).}
\label{tab:human_eval_scale}
\setlength{\tabcolsep}{5pt}
\begin{tabular}{ccc}
\toprule
\textbf{Agree. Rate} & \textbf{Kendall Tau} & \textbf{Spearman Coeff.} \\
\midrule
66.67 & 72.48 & 78.01 \\
\bottomrule
\end{tabular}
\end{table}

\section{Expert-Reviewed Case Study}
\label{app:expert_case}

We further examine a representative Physical Understanding case for severe convection over the northeastern United States on July 6, 2021. With Qwen3.7-Plus, \model-RAG raises the normalized score from 0.4 for \model-Base to 0.8. It retrieves six historical analogues, including four convective cases and five with overlapping regional conditions, then checks the analogy against target-time HRRR fields and meteorological diagnostics. As summarized below, the resulting explanation correctly connects a warm, moist, unstable environment and sufficient shear to organized convection, damaging downburst winds, and hail. A domain expert reviewed the complete input, observable tool-use trajectory, execution observations, and final answer, and confirmed that the evidence use and meteorological conclusion were professionally reasonable.

\begin{tcolorbox}[
enhanced,
breakable,
width=\linewidth,
colback=cyan!3,
colframe=teal!65!black,
colbacktitle=teal!15,
coltitle=black,
fonttitle=\bfseries\sffamily,
fontupper=\small\sffamily,
boxrule=0.7pt,
boxsep=3pt,
left=4pt,
right=4pt,
title={Representative \model-RAG Physical Understanding Case}
]
\textbf{Input.}
At 20:23 UTC on July 6, 2021, severe weather affected a corridor from eastern Pennsylvania and New Jersey through southern New England. Explain the meteorological setup and key forcing ingredients in one focused paragraph.

\textbf{Observable agent trajectory.}
\begin{enumerate}[leftmargin=*,itemsep=2pt,topsep=2pt]
    \item \textbf{Retrieve analogues.} The agent calls \texttt{access\_historical\_cases}, scans all 1,897 eligible cases, and ranks them by event type, region, and task form. It retains six cases, including prior convective-wind episodes over Pennsylvania, New Jersey, New York, and southern New England.
    \item \textbf{Inspect target-time conditions.} It calls \texttt{access\_hrrr\_reanalysis} and \texttt{access\_spc\_mesoscale\_images} for the northeastern United States near 20 UTC. The returned HRRR fields show a warm surface environment, strong low-level moisture, and locally enhanced winds, while unavailable SPC panels are explicitly recorded rather than inferred.
    \item \textbf{Diagnose storm organization.} It applies \texttt{compute\_layer\_profile\_metrics}, \texttt{compute\_wind\_diagnostics}, and \texttt{visualize\_gridded\_field}. The observations include 850-hPa moisture up to 15.4\,g\,kg$^{-1}$ and 850--500-hPa bulk shear with a 95th percentile of 17.4\,m\,s$^{-1}$.
    \item \textbf{Synthesize the evidence.} The agent combines the regional analogues with the target-specific diagnostics, rather than copying a historical answer, to identify organized multicell convection and its principal hazards.
\end{enumerate}

\textbf{Output.}
The agent reports a warm and moist environment with substantial instability and locally stronger shear near mesoscale boundaries. It concludes that organized thunderstorms can produce damaging convective wind gusts and hail while moving into southern New England, with gradual weakening near the more stable coastal environment.
\end{tcolorbox}

\section{Prompts}
\label{app:prompts}
This section presents the prompts used throughout \model. They include general prompts for individual-task execution, experience-grounding prompts for \model-RAG, \model-Skill, and \model-Modeling, operational-chain prompts for coordinating the sequential warning chain, and recovery prompts for handling execution, formatting, and timeout failures.

\tcbset{width=0.9\linewidth,colback=gray!5,colframe=black,coltitle=white,fonttitle=\bfseries,colbacktitle=black!75,boxsep=4pt,boxrule=0.6pt,left=3pt,right=3pt,breakable,fontupper=\ttfamily}

\subsection{General Prompts}

\begin{tcolorbox}[title=Individual Initial Prompt]
\small
You are an extreme-weather early-warning agent that solves the target task through evidence-grounded reasoning and interactive code execution.\\

\texttt{<WORKFLOW AND ACTION RULES>}\\

\texttt{<CODE EXECUTION AND RESPONSE RULES>}\\

Available multimodal evidence:\\
\texttt{<EVIDENCE SUMMARY>}\\

Available analytical tools:\\
\texttt{<TOOL DESCRIPTIONS>}\\

Target question:\\
\texttt{<TARGET QUESTION>}\\

Task-specific guidance:\\
\texttt{<TASK GUIDANCE>}\\

Experience-grounding guidance, when enabled:\\
\texttt{<EXPERIENCE GUIDANCE>}\\

Accumulated skill guidance, when available:\\
\texttt{<ACCUMULATED SKILL GUIDANCE>}\\

Prior execution observations:\\
\texttt{<EXECUTION HISTORY>}\\

Available intermediate artifacts:\\
\texttt{<ARTIFACT SUMMARY>}
\end{tcolorbox}

\begin{tcolorbox}[title=Workflow and Action Rules]
\small
Follow an iterative reason--execute--observe workflow.\\

1. At each turn, briefly state what is known and what evidence is still needed.\\
2. If further evidence is required, provide a short plan and pseudocode, then take exactly one code-execution action.\\
3. Use the returned observation to revise the next action. Do not repeat an unsuccessful action without addressing its failure.\\
4. When the task depends on external evidence or analytical tools, do not produce the final solution before at least one successful execution observation.\\
5. Finish only when the accumulated evidence supports the requested answer. Preserve the exact answer type and format required by the target question.\\
6. For an operational chain, solve only the current individual procedure and carry its completed answer forward before moving to the next procedure.
\end{tcolorbox}

\begin{tcolorbox}[title=Code Execution and Response Rules]
\small
Use only the documented analytical tools and the guarded code-execution interface. Do not access hidden storage paths or tool implementations.\\

Inside an execution action, define a callable \texttt{run(tools)}. Use code for data access, filtering, aggregation, statistical analysis, visualization, and task-specific computation. Return only compact textual findings and the images that must be inspected in the next reasoning turn. Keep large intermediate objects in reusable artifacts.\\

Valid execution response:\\
\texttt{<plan>}\\
A concise list of the remaining evidence-gathering steps.\\
\texttt{</plan>}\\
\texttt{<pseudocode>}\\
A compact description of the next computation.\\
\texttt{</pseudocode>}\\
\texttt{<execute>}\\
\texttt{def run(tools):}\\
\texttt{\   ...}\\
\texttt{</execute>}\\

Valid final response:\\
\texttt{<solution>}\\
The final answer in the format required by the target question.\\
\texttt{</solution>}\\

Each response must contain exactly one action: either one execution action or one final solution. Never place executable code outside the execution block or a final answer outside the solution block.
\end{tcolorbox}

\begin{tcolorbox}[title=Observation Continuation Prompt]
\small
Continue solving the same extreme-weather task using the newest execution result.\\

Latest observation:\\
\texttt{<LATEST OBSERVATION>}\\

Remaining objective:\\
\texttt{<REMAINING OBJECTIVE>}\\

Accumulated skill guidance, when available:\\
\texttt{<ACCUMULATED SKILL GUIDANCE>}\\

Relevant prior observations and artifacts:\\
\texttt{<EXECUTION HISTORY AND ARTIFACTS>}\\

Treat the latest observation as the newest evidence. Either gather the next necessary piece of evidence through one execution action or return the final solution when the evidence is sufficient.
\end{tcolorbox}

\subsection{Experience-Grounding Prompts}

\begin{tcolorbox}[title=Experience Guidance: \model-RAG]
\small
You are the case-retrieval stage of \model-RAG. Use historical cases as analogical evidence for the target task.\\

Target question and event conditions:\\
\texttt{<TARGET TASK>}\\

Historical case collection:\\
\texttt{<HISTORICAL CASES>}\\

1. Inspect the complete historical collection relevant to the target subtask.\\
2. Rank cases by their usefulness for the target task. Consider task intent, event conditions, reference time, location, event type, question form, answer format, source variables, and available labels.\\
3. Select at most six cases. Preserve each selected case's question, reference answer, and event metadata so that its analogy remains interpretable.\\
4. Use the selected cases as evidence and answer-format references. Do not copy a historical answer without checking it against the current event evidence.\\
5. Continue the evidence-grounded solving workflow with the selected cases in context.
\end{tcolorbox}

\begin{tcolorbox}[title=Experience Guidance: \model-Skill Rehearsal]
\small
You are the rehearsal stage of \model-Skill. Historical cases are practice tasks for acquiring reusable solving procedures, not direct answer examples.\\

Target question and event conditions:\\
\texttt{<TARGET TASK>}\\

Historical case collection:\\
\texttt{<HISTORICAL CASES>}\\

Current skill guidance:\\
\texttt{<CURRENT SKILL GUIDANCE>}\\

1. Inspect the complete historical collection relevant to the target subtask and select at most three cases with the highest rehearsal value.\\
2. For each selected case, hide its reference answer and solve the rehearsal question using the same evidence-grounded workflow as the target task.\\
3. After completing the rehearsal, reveal the reference answer and invoke the Skill Refinement Prompt to compare the solution process with the reference.\\
4. Retain only reusable procedural guidance. Do not store event-specific answers or details that would not transfer to another case.\\
5. Repeat rehearsal and refinement for the selected cases, then solve the target task using the accumulated skill guidance.
\end{tcolorbox}

\begin{tcolorbox}[title=Skill Refinement Prompt]
\small
You are refining the reusable skill guidance of \model-Skill after one rehearsal case.\\

Existing skill guidance:\\
\texttt{<CURRENT SKILL GUIDANCE>}\\

Rehearsal question and event conditions:\\
\texttt{<REHEARSAL TASK>}\\

Rehearsal solution and execution observations:\\
\texttt{<REHEARSAL TRAJECTORY>}\\

Reference answer:\\
\texttt{<REFERENCE ANSWER>}\\

1. Identify which reasoning steps, evidence choices, tool-use strategies, or answer-format decisions were effective.\\
2. Diagnose errors by comparing the rehearsal solution with the reference answer and the supporting evidence.\\
3. Add or revise only short, actionable guidance that can improve future tasks of the same kind. Keep one reusable procedure per item.\\
4. Preserve useful existing guidance and remove an item only when the rehearsal provides clear evidence that it is misleading.\\
5. Exclude the rehearsal's final answer and event-specific facts from the refined skill.\\

Return only the revised skill guidance.
\end{tcolorbox}

\begin{tcolorbox}[title=Experience Guidance: \model-Modeling]
\small
You are the modeling stage of \model-Modeling. Convert the historical case collection into a trained task-specific predictor that provides predictive evidence for the target task.\\

Target question and event conditions:\\
\texttt{<TARGET TASK>}\\

Historical case collection:\\
\texttt{<HISTORICAL CASES>}\\

Available ML development tools:\\
\texttt{<ML TOOLS>}\\

1. Use the complete historical collection relevant to the target subtask. Construct explicit training and validation samples with task-appropriate inputs and targets.\\
2. Design features from historical cases and available meteorological, spatial, temporal, or impact evidence. Prevent target leakage.\\
3. Select an appropriate ML model, train it on the training split, and evaluate it on held-out validation samples with a task-appropriate metric.\\
4. If the requested answer is not directly learnable, define a learnable intermediate target that can be mapped to the required answer, then train and validate a model for that target.\\
5. Apply the trained predictor to the current task and return its prediction, validation evidence, and the information needed to interpret the prediction.\\
6. Do not replace model training with nearest-neighbor retrieval, descriptive statistics, or heuristic scoring alone. Continue to the target-solving phase only after a trained and validated predictor is available.
\end{tcolorbox}

\subsection{End-to-end Operational-Chain Prompts}

\begin{tcolorbox}[title=Operational-Chain Initial Prompt]
\small
You are solving an operational extreme-weather warning chain in one continuous conversation. Work on exactly one individual procedure at a time while preserving the shared event context.\\

Full operational-chain question:\\
\texttt{<FULL CHAIN QUESTION>}\\

Individual procedures and required answer formats:\\
\texttt{<PROCEDURE DEFINITIONS>}\\

Current individual procedure:\\
\texttt{<CURRENT PROCEDURE>}\\

Available evidence and analytical tools:\\
\texttt{<EVIDENCE AND TOOLS>}\\

Task-specific and experience-grounding guidance:\\
\texttt{<TASK AND EXPERIENCE GUIDANCE>}\\

\texttt{<WORKFLOW AND ACTION RULES>}\\

\texttt{<CODE EXECUTION AND RESPONSE RULES>}\\

Solve only the current individual procedure. Return its answer in the required format and do not assemble the final chain response until every procedure has been completed.
\end{tcolorbox}

\begin{tcolorbox}[title=Operational-Chain Transition Prompt]
\small
The previous individual procedure is complete. Continue the same operational chain with the shared event context and prior answers in scope.\\

Completed procedure answers:\\
\texttt{<COMPLETED PROCEDURE ANSWERS>}\\

Next individual procedure:\\
\texttt{<NEXT PROCEDURE>}\\

Guidance for the next procedure:\\
\texttt{<NEXT-PROCEDURE GUIDANCE>}\\

Use historical experience only for the current individual procedure. Solve the next procedure in its required answer format, preserve the completed answers, and defer final chain assembly until all procedures are complete.
\end{tcolorbox}

\subsection{Recovery Prompts}

\begin{tcolorbox}[title=Execution Recovery Prompt]
\small
The previous code execution failed. Correct the failure before continuing the task.\\

Failed action and returned error:\\
\texttt{<FAILED EXECUTION>}\\

Earlier unsuccessful attempts:\\
\texttt{<RECOVERY HISTORY>}\\

Remaining objective:\\
\texttt{<REMAINING OBJECTIVE>}\\

Identify the likely cause, revise the plan, and produce one corrected execution action. Use documented tools only, repair invalid arguments or result structures, and change strategy when the same failure has already occurred. Do not return a final solution in this recovery turn.
\end{tcolorbox}

\begin{tcolorbox}[title=Output-Format Recovery Prompt]
\small
The previous response did not follow the required output format. Correct its format without redoing the analysis or introducing new facts.\\

Required answer or action format:\\
\texttt{<REQUIRED FORMAT>}\\

Previous invalid response:\\
\texttt{<INVALID RESPONSE>}\\

Format error:\\
\texttt{<FORMAT ERROR>}\\

Return one corrected response that preserves the intended content and satisfies the required format exactly.
\end{tcolorbox}

\begin{tcolorbox}[title=Timeout Recovery Prompt]
\small
The previous code execution exceeded the time limit. Produce a faster evidence-gathering action.\\

Timed-out action:\\
\texttt{<TIMED-OUT EXECUTION>}\\

Evidence already available:\\
\texttt{<AVAILABLE OBSERVATIONS>}\\

Remaining objective:\\
\texttt{<REMAINING OBJECTIVE>}\\

Reduce the scope of the next computation, avoid repeated work and broad scans, and reuse existing observations or artifacts. Return one revised execution action and do not return a final solution in this recovery turn.
\end{tcolorbox}

\end{document}